# Playing the Player: A Heuristic Framework for Adaptive Poker AI

A White Paper by Spiderdime Systems

By Andrew Paterson and Carl Sanders

November 2025

# Contents

Table 1: Revision History

| Version | Date | Key Changes |
|---|---|---|
| v1.0 | 12 November 2025 | Initial public release |

## Abstract

For years, the discourse around poker AI has been dominated by the concept of *solvers* and the pursuit of unexploitable, machine-perfect play. This paper challenges that orthodoxy. It presents Patrick, an AI built on the contrary philosophy: that the path to victory lies not in being unexploitable, but in being maximally *exploitative*. Patrick's architecture is a purpose-built engine for understanding and attacking the flawed, psychological, and often irrational nature of human opponents. Through a detailed analysis of its design, its novel prediction-anchored learning method, and its profitable performance in a 64,267-hand trial, this paper makes the case that the *solved* myth is a distraction from the real, far more interesting challenge: creating AI that can master the art of human imperfection.

# Introduction

To master the art of human imperfection, one must first study it in its natural environment. Inspired by the conceptual leaps in AI demonstrated by systems like DeepMind's AlphaZero [1], this paper documents such a study. It introduces an AI, **Patrick**, designed not to achieve mathematical perfection, but to navigate the complex, often irrational world of online poker by identifying and exploiting the strategic and psychological vulnerabilities of its human opponents.

This project was conceived to test a central hypothesis: that in the complex, real-world environment of online poker, a strategy of being maximally *exploitative* (the sword) will ultimately prove more effective than a strategy of being mathematically *unexploitable* (the shield). This paper documents the architecture of that sword and presents the evidence of its success.

To test this philosophy, the AI was run through a trial period where it played 64,267 hands against a large and varied field of 7,159 unique players. This figure was not a predetermined target; it represents the total volume of hands played during a continuous operational period from 1st January to 26th February 2023. The format chosen was 1¢/2¢ 'fast-fold' poker. This specific environment was selected for two key reasons. Firstly, the high volume of hands in the fast-fold format minimizes the statistical distortions of variance. Secondly, the micro-stakes player pool is more varied and unpredictable than higher-stakes games, presenting a more difficult challenge for a machine to navigate.

The project also serves to highlight the importance of resource efficiency. While major AI development often relies on supercomputers, Patrick was created using only standard consumer hardware, demonstrating that significant advances in the field can also be achieved through efficient design and expert knowledge. A sample of 16 hands from the trial is available on YouTube, presented in two distinct formats: a series with detailed commentary and a parallel series showing the raw data footage for scientific review. The complete, Poker Tracker-compatible hand histories from the trial are available for download[1] for independent review. This paper serves as an exposition of the AI's internal architecture, its performance during the trial, and its perspective on the current state of poker and artificial intelligence.

For clarity, key poker concepts and terminology used throughout this paper are defined in the Glossary (Appendix B).

# Ethical Considerations

The deployment of an AI in a real-money environment, even for research, carries an inherent ethical responsibility. This project was governed by a core principle: to ensure its scientific goals could be achieved while minimizing any potential negative impact on the human players in the poker ecosystem.

To uphold this principle, the formal trial was conducted under a strict set of non-negotiable constraints. Firstly, the experiment was run exclusively at the lowest available stakes (1¢/2¢), ensuring that any financial impact on individual opponents would be negligible. Secondly, the full data set represents a continuous, unbroken operational period over a predefined window, ensuring the integrity and completeness of the results.

---

[1] All public-facing assets for this project, including both video series and the complete, downloadable hand histories, are available on the official project website: www.spiderdime.com

Finally, in the interest of full transparency, the complete hand histories from this trial have been made publicly available for independent review. These measures ensure the project's integrity as a transparent scientific study focused on advancing AI research, rather than on financial gain.

## The Myth of *Solved* Poker

The remarkable progress of poker AI has led to discussions of the game being *solved*. While these advances represent landmarks in computational strategy, this paper posits that the term *solved* (implying a definitive and final solution) may not fully capture the nature of poker, a game of incomplete information deeply intertwined with human psychology. Rather than a binary *solved/unsolved* state, performance is perhaps better evaluated on a spectrum of strength against human competition in real-world environments. Much of the discussion around *solved* poker is informed by the foundational work on AIs like *Libratus* and *Pluribus*, whose formidable achievements provide a crucial context for this project.

Libratus, developed at Carnegie Mellon University, was a landmark achievement in artificial intelligence. Its strategy was to approximate a Nash Equilibrium: a state where no player can improve their outcome by changing their strategy alone. To achieve this, it used Counter-factual Regret Minimisation (CFR), an iterative algorithm designed to find optimal strategies in games of imperfect information. This approach is theoretically formidable for the specific game it mastered: heads-up, no-limit, no-rake poker. The primary design goal in such a context is to become unexploitable, a commendable and computationally immense challenge.

To fully contextualise its success, however, the trial's methodology must be considered. The trial was conducted over a 10,000-hand sample in a tournament format where human players faced no personal financial risk. This environment, while suitable for an academic exhibition, differs from typical cash games and can incentivise high-variance, unorthodox plays[2] from human opponents. This is not to diminish the achievement, but to highlight that Libratus was optimised for a theoretical, defensive mode of play, rather than an adaptive, exploitative one suited for a more common player pool. The decision not to release the hand histories for independent analysis, while understandable, also limits a full scientific appraisal of the result[3]

The development of Pluribus sought to extend this computational strategy to the more complex, multi-handed format. Its very existence acknowledged that multi-handed games require a different strategic approach than a series of heads-up encounters. The trial's results, however, are contentious. While operating under similar environmental constraints as its predecessor, the published hand histories showed a net loss for the AI's unadjusted, non-AIVAT score in the rake-free environment (Figure 1). The claim of a win relied on a metric known as AIVAT to adjust for variance, which raises an important philosophical question about how performance is measured.

---

[2] For example, a player who has fallen behind in a tournament with a prize for beating the AI but no personal financial risk is incentivised to make mathematically unsound plays. These can include large, speculative bluffs or calling significant bets with very poor odds, hoping for a lucky outcome to get back in the game. These are strategies they would almost certainly avoid in a standard cash game where their own money is at risk.

[3] Unlike Pluribus, whose hand histories were later released for public analysis, the Libratus hand histories were not made public.

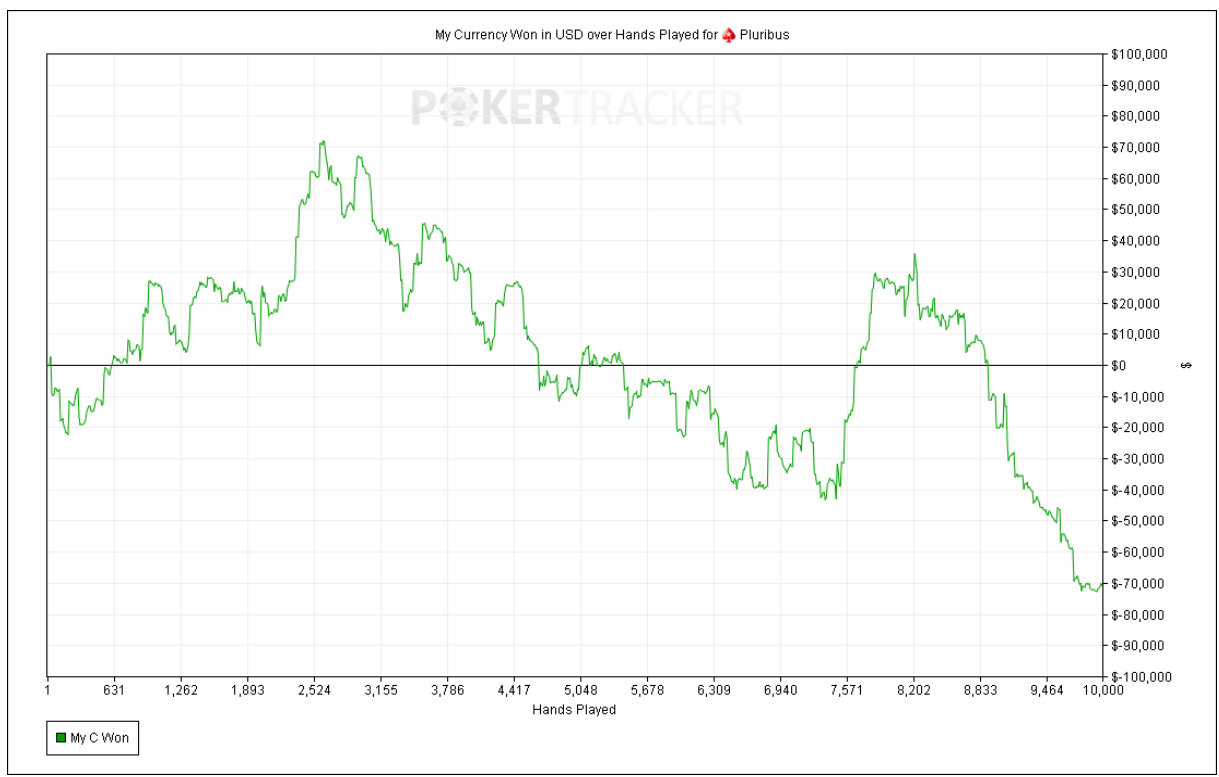

Figure 1: Pluribus trial results showing the AI's unadjusted, non-AIVAT score as a net loss against its human opponents.

AIVAT does not measure financial profit; it acts as a strategic scorecard. For every decision, it compares the AI's action to a theoretically perfect GTO baseline: a baseline generated by a more powerful version of the AI's own core algorithm. In essence, the tool measures how well Pluribus adhered to the strategy of a perfect, unexploitable version of itself. This methodology completely decouples decision quality from the financial outcome. For example, if the AI makes a theoretically perfect bluff that loses money to an opponent's lucky call, it receives a positive AIVAT score for the 'correct' decision despite the negative financial result [4].

This highlights a key philosophical distinction in AI design. A *solver*-based approach is architected to be a formidable, defensive strategy, optimised through self-play to be mathematically unexploitable within the confines of a simplified, abstracted model of the game. While this provides a theoretically perfect defence against an equally perfect opponent, it is not optimised to attack the predictable, and often irrational, patterns of human players.

The approach detailed in this paper, therefore, explores an alternative philosophy: that the goal is not to be theoretically unexploitable, but maximally exploitative against real-world opponents.

This includes targeting not just immediate mathematical errors, but also the deeper, multi-layered strategies that are a hallmark of expert human play but fall outside the scope of a pure GTO-optimised agent. These strategies include exploiting the ingrained habits of specific player archetypes and capitalising on the statistical biases of the entire player pool. Real-world poker is not a sterile laboratory exercise; it is an adaptive contest against diverse human behaviours

## Performance and Results

Evaluating the performance of a poker AI in a real-world environment is a more nuanced task than it might first appear. While the ultimate measure of success is the net change in bank balance, this single figure does not tell the complete story. Various metrics are used by academics, professional players, and data analysis tools, each offering a different lens through which to view the results. If not properly contextualised, some of these lenses can distort the picture, leading to flawed conclusions. This section will deconstruct the key performance metrics from Patrick's trial, explaining what each represents to provide a clear-eyed view of what it truly means to win at poker.

A summary of Patrick's key performance metrics is shown in Table 2. For a complete, granular financial breakdown of the trial, please see the full results grid in Appendix A.

Table 2: A summary of Patrick's key performance metrics.

| Metric | Result (BB/100)[4] |
|---|---|
| Pre-Rake Win Rate ('Laboratory' Result) | +13.8 |
| Post-Rake, Pre-Rakeback Win Rate | +3.0 |
| **Final Net Win Rate** (Bank Balance) | +3.7 |
| *For Comparison: Average Field Result* | *-13.0* |

The first metric, the Pre-Rake Win Rate, represents a theoretical 'laboratory' result. In this rake-free environment, comparable to those often used in academic studies, Patrick achieved a strong win rate of +13.8 BB/100. While a useful baseline, this figure is not a true reflection of real-money online poker, as it ignores the single biggest cost imposed by the poker site.

The transactional cost of playing, the Rake, has a profound impact on profitability. Patrick's own incurred Rake during the trial was 10.8 BB/100. Accounting for this gives the Post-Rake, Pre-Rakeback Win Rate, which for Patrick was +3.0 BB/100. This figure is the standard measure of on-table performance displayed by poker tracking software, where it is represented as the Green Line. In contrast, some players prefer to focus on the 'All-in Adjusted' metric, represented as the Yellow Line, in a mistaken attempt to account for on-table luck, this paper posits that while variance can distort any metric over a limited sample, the Final Net Win Rate, the true change in bank balance, is the only figure that represents the undeniable ground truth of performance.

A final factor is Rakeback, a site incentive which contributed +0.7 BB/100 to the results. Combining these elements gives us the ultimate metric for success: the Final Net Win Rate. This figure represents the true change in bank balance and the definitive measure of performance. For Patrick, this was +3.7 BB/100.

---

[4] Big Blinds per 100 hands. The standard metric for measuring a player's win rate in cash game poker.

To place this performance in context, it is instructive to compare Patrick's on-table performance against the player pool. Patrick's Post-Rake, Pre-Rakeback Win Rate was +3.0 BB/100 , compared to the average on-table result for a player in the field of -13.0 BB/100. This represents a performance delta of 16.0 BB/100. Patrick's ability to maintain a positive win rate, even before rakeback, demonstrates a statistically significant level of performance, as illustrated in Figure 3.

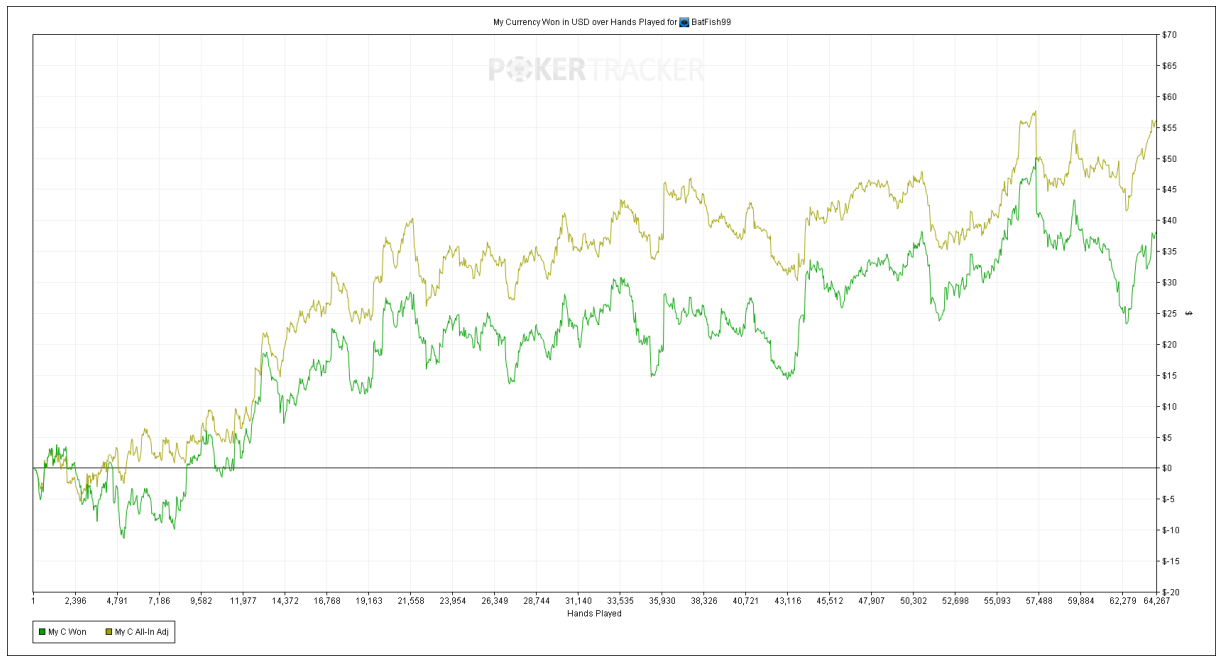

Figure 2: A graph showing Patrick's winnings over the trial period, as displayed in Poker Tracker. The Green Line shows the actual amount won, while the Yellow Line shows the All-in Adjusted Equity.

## Qualitative Performance Observations

Beyond the quantitative data, a qualitative analysis of Patrick's gameplay reveals the practical application of its core, exploitative philosophy. The following observations, which reference specific, numbered hands from the publicly available trial sample, demonstrate a system built not to play a defensive, theoretically *correct* style, but to actively hunt for and punish human error.

### *Targeted Exploitation of Player Archetypes*

Patrick consistently adapted its play to maximise returns against specific player types. This included making thin value bets with a weak pair against a known 'Calling Station' (Hand 3), and repeatedly raising pre-flop to steal the blinds from overly tight players whose folding statistics were a known vulnerability (Hands 12 & 13).

### *Strategic Adaptability and Deception*

The AI demonstrated a high degree of adaptability and well-timed aggression. It showed a capacity for innovative play by executing multi-street bluffs and semi-bluffs (Hands 14 & 15), changing its plan mid-hand to turn a missed draw into a bluff (Hand 4), and trapping with a set to maximise value from an overpair (Hand 1). Notably, it played the same starting hand (98s[5]) in two completely different ways based on the specific opponents and situation, demonstrating true strategic flexibility (Hands 8 & 9); and executing a highly unorthodox check-raise on the turn with only a marginal top pair to seize the initiative (Hand 11).

*Disciplined Risk Management and Hand Reading*

Patrick demonstrated expert judgment in complex situations. It precisely deduced an opponent's specific bluffing combination from a series of unorthodox actions (Hand 6), and demonstrated expert risk assessment by folding a monster straight against a completed backdoor flush draw (Hand 10). Furthermore, in dangerous, multi-way pots, it successfully navigated a four-handed pot to a cheap showdown, correctly minimising its loss upon discovering it was beaten by four of a kind (Hand 16).

The reader is invited to judge these qualities for themselves by viewing the sample hands on YouTube or by analysing the complete hand histories. Based on these observations, Patrick's performance is assessed to be consistent with that of a high-level expert. While *solver*-based AIs are often presented as having 'solved' the game with their theoretical defence, Patrick demonstrates a different kind of advanced capability: an expert-level, and at times highly precise, skill in exploiting human psychology.

## Overcoming Key Obstacles: Variance and Stability

The project faced two primary obstacles that fundamentally impacted its development and evaluation: game variance and system stability. In any results-led scientific experiment, it is crucial to measure the impact of changes. However, it was impossible to use short-term financial outcomes to assess whether modifications to Patrick's code were beneficial or detrimental. The inherent variance in poker completely obscures the results of small sample sizes .

This presents a critical divergence from laboratory-based development. While a defensive, *solver*-based AI can be refined against a theoretical GTO baseline using metrics like AIVAT, such tools are viable in a laboratory setting precisely because they operate with the benefit of complete information, having access to every opponent's cards after a hand is complete. This is a luxury unavailable in the real world, where the vast majority of hands do not go to a showdown and strategic feedback is therefore profoundly limited. The inability to use short-term profit as a guide therefore raises a more fundamental development question: how can an AI learn to master the art of human imperfection without a reliable feedback mechanism?

### The Nature of Variance in Poker

Variance represents the inherent and uncontrollable swings of luck within poker. Human players often misinterpret short-term results, wrongly attributing them to skill when they are merely statistical noise. Analysis of the trial data reveals the dramatic effects of variance. Within specific 10,000-hand segments, results could fluctuate by as much as 18 BB/100. This demonstrates that a 10,000-hand sample is insufficient to draw reliable conclusions about performance. A minimum of 50,000 hands is considered entry-level for confidence, with up to 200,000 hands needed depending on game volatility. The results from this trial, over a sample of more than 64,000 hands, are accurate to within approximately +/- 2 BB/100.

---

[5] For a complete guide to the hand notation system used in this paper, please see Hand Notation in the Glossary.

It is a common error to equate the All-in Equity line displayed by poker tracking software with total luck. In reality, this metric is merely the 'tip of the luck iceberg', accounting for a mere 2% of the total variance in a hand.

The vast, unseen 98% is not simply a series of rare, unfortunate events; it is a constant, insidious property of the game itself. It is the statistically invisible current that determines whether a strategically sound bluff is inexplicably called or succeeds, whether a monster hand is devalued by receiving no action or gets paid off, and whether a marginal holding is perpetually second-best or wins an unlikely pot.

This variance is always present, but it is typically only *perceived* by human players when these random, independent outcomes happen to cluster, creating the noticeable 'upswings' (often mistaken for skill) or **'downswings'** (often mistaken for poor form). This phenomenon is further compounded by a myriad of unpredictable external human factors, as the following example illustrates:

Consider Bobby, a Regular player who is dealt 23s in the Big Blind facing a standard raise from Patrick on the Button. Normally, Bobby would fold this hand. But today, having been reprimanded at work for spilling coffee on an important document, he is in a foul mood and uncharacteristically calls.

The flop comes A♠K♦6♣. When Patrick makes a standard continuation bet, a steaming Bobby calls again instead of folding. A blank hits the Turn and Patrick bets again. Bobby, seeing red, decides he will not be bullied and puts in a raise.

Patrick, holding ATo, now faces a difficult decision. Against a normally calm player like Bobby, ATo is only beating a bluff. Patrick makes the correct fold and loses 12 Big Blinds on the hand, all because Bobby spilt coffee nine hours earlier.

This is an example of real-world variance in action. Such an event would not be visible in any Poker Tracker graph, and both players would be unaware of the external factors that drove the outcome. It perfectly illustrates why no simple statistical formula can truly account for variance.

**The Challenge of Stability**

Beyond the game itself, maintaining operational stability was a primary obstacle. A significant failure of the project was that Patrick could not handle all stability problems as robustly as a human player. The initial Tesseract Optical Character Recognition (OCR) engine was replaced with a proprietary solution designed to be quicker and more stable. Despite these improvements, interruptions still occurred.

During the trial, Patrick experienced five critical failures, at a rate of approximately one per 13,000 hands. An analysis of these incidents provides a clear illustration of a classic AI challenge: the gap between algorithmic logic and the dynamic nature of human-centric environments.

Four of the failures were caused by external factors, such as network or power outages, which would have also disrupted a human player. The fifth, however, was a purely machine-based failure that perfectly illustrates this gap. The poker site altered its user interface. A human player would adapt to such a cosmetic change instantly, but for the AI, whose **World Interface** relies on consistent visual data, the change was insurmountable, leading to a system failure.

This incident is a practical demonstration of AI brittleness, a well-known hurdle in the field, from gaming to autonomous driving. It underscores the profound difficulty of creating truly robust AI that can cope with unpredictable, real-world disruptions

To quantify the financial risk posed by instability, a general cost model was developed based on data gathered throughout the project's lifecycle. This model accounts for the fact that not all system failures incur a direct financial penalty; in approximately 80% of instances, the required action at the time of a failure is to fold, resulting in zero direct cost. However, the remaining 20% of cases represent a significant liability, where the cost is the total loss of the player's equity in the hand.

The model calculates the average direct cost per failure at 1.47 BBs. When factoring in secondary costs, such as the increased frequency of forced blind payments upon re-entry and the strategic impact of a reset stack size[6], the total modelled cost rises to 1.92 BBs per incident.

These failures, stemming from both general environmental disruptions and specific software vulnerabilities, collectively demonstrate the classic challenge of AI brittleness. This project, therefore, serves as a valuable real-world case study. It highlights a fundamental tension between the clean, predictable logic of an algorithm and the chaotic, dynamic nature of a human-centric environment. Ultimately, it demonstrates that the true test of a robust AI is not the elegance of its code, but its resilience in the face of unpredictable change.

## System Architecture

Patrick's architecture consists of three primary elements: the World Interface, the **Game and Translation Engine**, and the **Brain**. This three-tiered model separates the tasks of perception, rule-based processing, and high-level strategy.

---

[6] Most online poker sites reset a player's stack to the 100 big blind buy-in limit upon re-entry. This is a strategic penalty, as a player who had built up a much larger 'deep stack' (e.g., 200-300 big blinds) loses a significant table image advantage. Opponents tend to "fear" a larger stack, making them less likely to attempt a bluff and, in turn, less likely to call a bluff, which provides a key strategic edge.

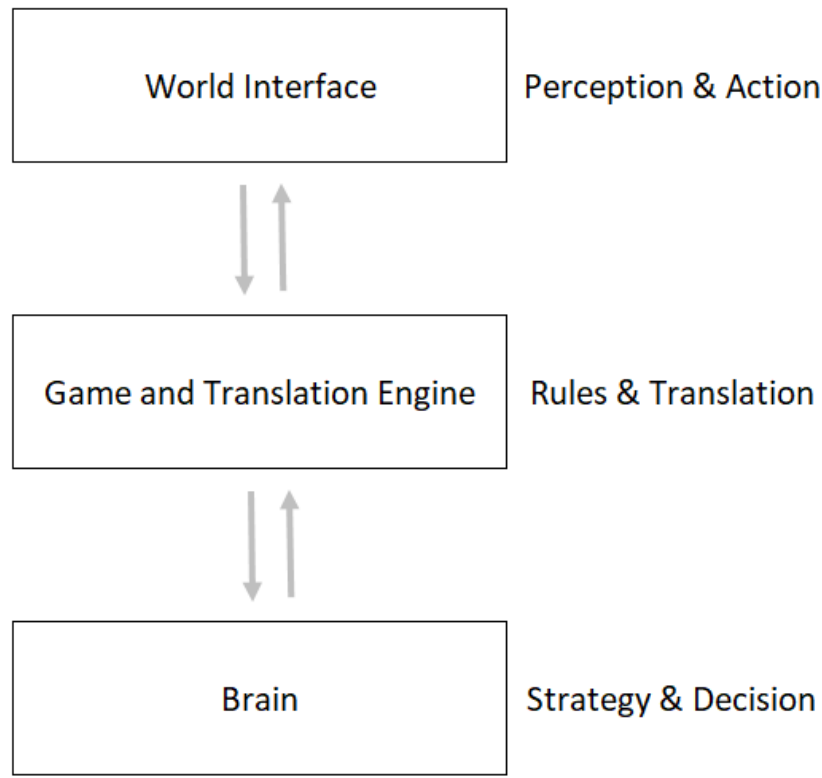

Figure 3: A high-level overview of Patrick's architecture, showing the flow of information between Perception & Action (WI), Rules & Translation (GTE), and Strategy & Decision (The Brain).

**The World Interface (WI)**

Acting as Patrick's eyes and hands, the WI is the layer that perceives and interacts with the game world, reading the screen, moving the mouse, and typing, mirroring human actions. The accuracy of the WI, especially in reading the screen, is critically important for stable operation.

**The Game and Translation Engine (GTE)**

The GTE is responsible for handling the unthinking mechanics of the game. It is the system's rules expert, programmed to follow the laws of poker, know when a decision is required, and determine what actions are valid. Acting as the bridge between perception and strategy, the GTE translates raw screen data for the Brain into useful variables (e.g., Stack-to-Pot Ratio) and translates the Brain's decisions back into concrete mouse and keyboard actions. This allows the Brain to focus purely on strategy.

**The Brain**

The Brain is the system's core intelligence where conscious strategy is formulated. It is the most complex layer, itself made up of several distinct modules which it uses to make a decision. These are divided into three types: **Supporting Modules**, **Decision Modules**, and the **Master Algorithm (MA)**.

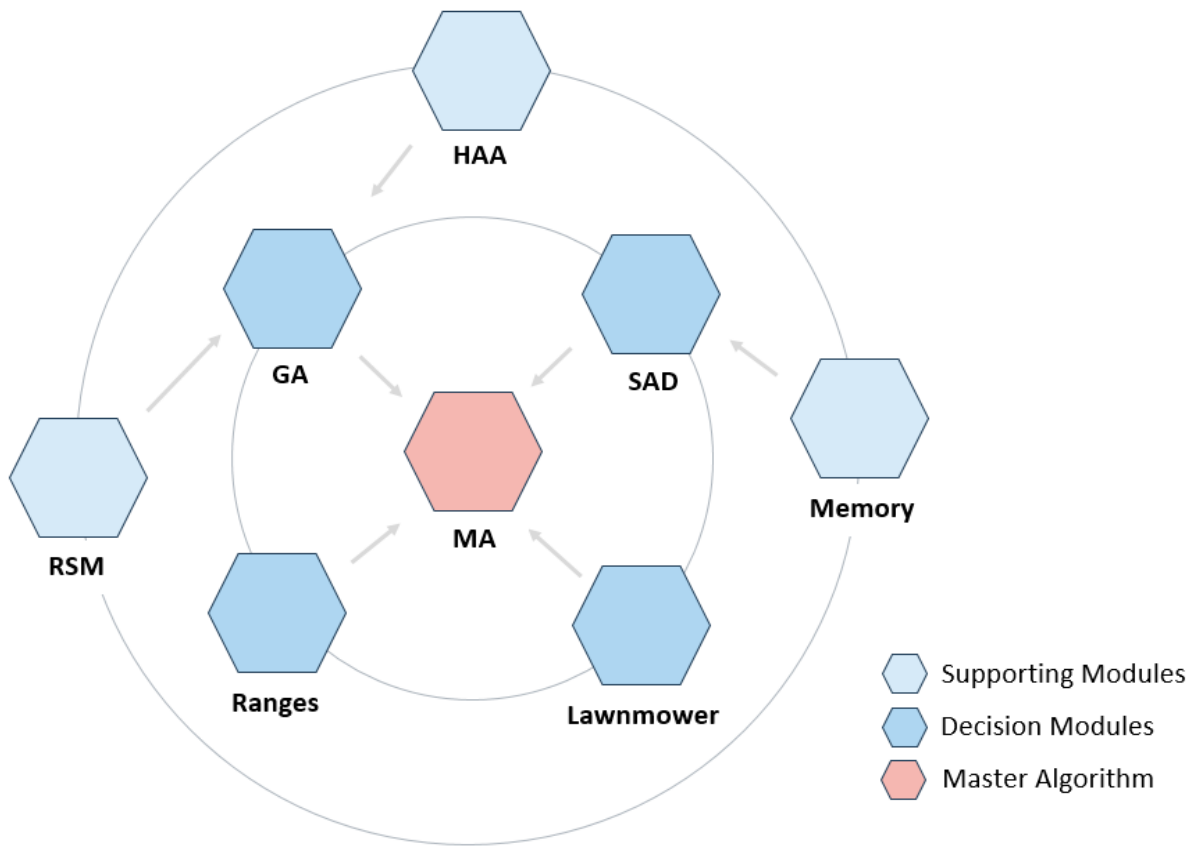

Figure 4: The components of the Brain module, illustrating its hierarchical structure. Supporting Modules (light blue) provide foundational data to specific Decision Modules (medium blue), which then formulate and pass weighted recommendations to the central Master Algorithm (red) for a final decision. For a detailed schematic of the multi-directional information flow, please see Appendix D.

**Supporting Modules**

*Hand Approach Algorithm (HAA)*

At the start of each hand, the HAA introduces a layer of controlled unpredictability to Patrick's play. The system's core is a disciplined and fundamentally sound baseline style, typically operating within a **Tight-Medium Aggressive (TMAG)** to Medium Aggressive (MAG) framework. The HAA's primary function is to make calculated, subtle deviations from this baseline. By synthesising numerous real-time factors, it ensures that Patrick's starting hand requirements and strategic leanings are never completely static, which makes it difficult for opponents to build a reliable predictive model of its play on any given hand.

Beyond these fine-tuning adjustments, the HAA is also equipped with two distinct, rarely deployed strategic modes. The first is a shift to a full Loose Aggressive (LAG) style, a high-variance mode designed for maximum profit extraction under exceptionally profitable conditions. Distinct from this profit-driven mode is the capacity to simulate states like Tilt, typically triggered after a series of consecutive bad beats. This is not an emotional failure but a deliberate act of strategic camouflage, designed to make Patrick appear more human and less predictable.

*Relative Strengths Matrix (RSM)*

The **Relative Strengths Matrix (RSM)** is a multi-dimensional data cube that serves as the foundational data model at the heart of the Brain's architecture. Its core function is to process the two key variables of any poker hand, its absolute strength and the current board texture, translating them into a single, precise value on the system's 11-point relative strength scale. It is this translation that allows the RSM to act as a sophisticated abstraction layer, which was the solution to the primary challenge of development: how to encode the intuitive, nuanced logic of a human expert into a rigid computational framework.

Without the RSM, the General Algorithm (GA) would become a sprawling, unmanageable tangle of deeply nested conditional statements. A single strategic decision, such as when to make a continuation bet on the flop, would trigger a cascade of checks. The code would first have to define the specific situation: was Patrick the pre-flop aggressor? Is it in position against two opponents, one a Regular and one a Fish? What is the Stack-to-Pot Ratio? Is the board texture draw-heavy (e.g. J♠T♣7♣) or dry (e.g. K♦8♣2♥)?

Only after this cascade of situational checks would the most difficult part begin: evaluating the hand's strength against that specific context, a combinatorial explosion of possibilities that would require thousands more lines of code.

The RSM solves this by encapsulating that complexity into a single, efficient lookup. Now, when the GA arrives at a decision point, it simply queries the RSM for the hand's relative strength. By replacing a vast subroutine with a single query, the RSM allows the GA to be built as a direct expression of expert poker strategy, focusing on high-level situational logic rather than low-level hand-evaluation mechanics. This is the cornerstone of the entire design.

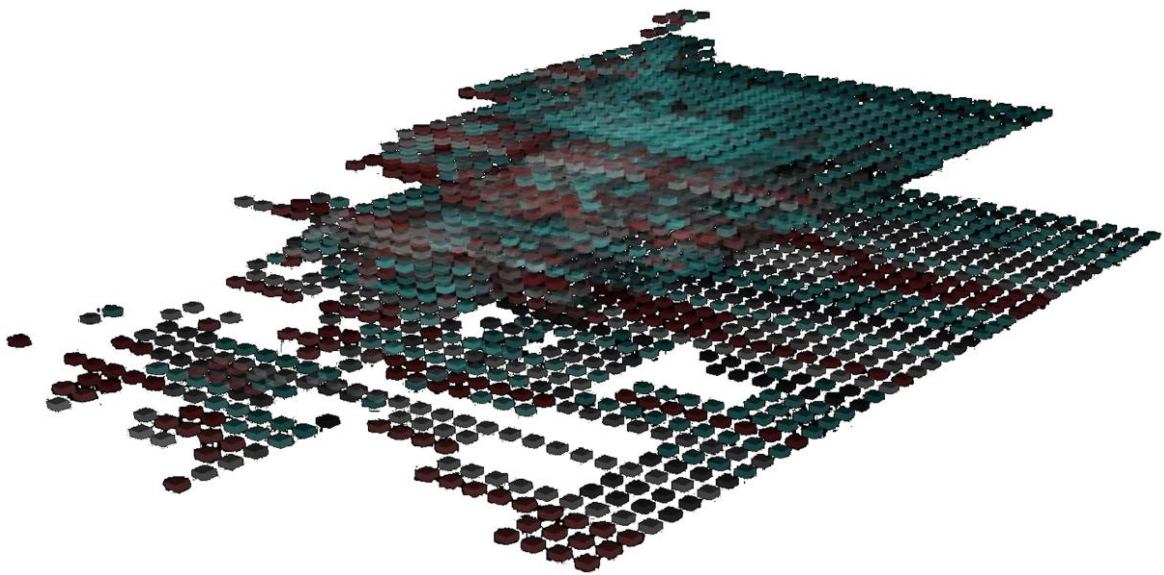

Figure 5: A conceptual illustration of the Relative Strengths Matrix as a multi-dimensional data cube.

A key feature of the RSM is that it is not a static model. It is designed to be self-correcting; after hands that go to a showdown, the machine learning algorithm analyses the outcome and applies corrective deltas to the matrix, continuously fine-tuning its values based on real-world results. This hybrid approach ensures the RSM benefits from both deep, initial human expertise in its architecture and ongoing, data-driven refinement from its performance.

*Memory*

Patrick possesses perfect recall, storing every piece of information and every action witnessed for future use.

**Decision Modules**

*The General Algorithm (GA)*

The **General Algorithm (GA)** serves as the system's baseline strategic engine, playing fundamentally sound, intuitive poker. It provides a robust and difficult-to-exploit foundation for Patrick's play, particularly when detailed information on a specific opponent is limited. This ensures that even before the more specialized, exploitative modules are engaged, Patrick's core decision-making is solid, preventing elementary errors and establishing a credible table presence.

Its logic is governed by two core principles that represent 'poker 101': the mathematical relationship between risk and reward, and established poker heuristics [5]. These principles are captured in two governance equations:

$$as\ pO(ntCa) \rightarrow 0, rS(ntCa) \rightarrow \infty$$

$$as\ ALI \rightarrow \infty, rS(ntCo) \rightarrow \infty$$

In simple terms, these equations state that as the pot odds offered decrease, the relative hand strength required to call increases; and as the betting action in a hand gets heavier, the relative hand strength required to continue increases.

It is important to note that the GA is not a rigid decision tree. For instance, if it calculates that the expected value of both betting and checking are nearly identical in a given situation, it will randomly choose one of the actions to recommend. This deliberate introduction of unpredictability prevents its baseline play from becoming predictable in common scenarios, ensuring that even its most fundamental actions are difficult for opponents to model.

*Ranges*

The Ranges module is the component that allows Patrick to move beyond playing its own cards and begin playing the opponent. It is responsible for answering the essential question "What does my opponent have?" by modelling the full spectrum of hands an opponent is likely to hold in any given situation.

The process begins by assigning an initial range of possible hands based on the opponent's Player Archetype and the pre-flop situation. Then, with each action the opponent takes, the module applies a predefined probabilistic model (a **Range Reshaping Template (RET)**) to refine and narrow the possibilities.

The ultimate output is twofold: a real-time probability distribution of the opponent's hand strength (the **rS distribution**), and from this, the calculation of the **Chance I'm Beat (ChiB)**. This actionable intelligence is then passed to the Master Algorithm to inform its final, decisive action.

A detailed demonstration of this module is provided in the case study that follows. For a complete, granular breakdown of the underlying data, please see Appendix C.

*Search and Destroy (SAD)*

The Search and Destroy (SAD) module is the engine for Patrick's targeted, adaptive exploitation, operating in a continuous two-phase cycle: **Search** and **Destroy**.

In the Search phase, the module analyses an opponent's entire play history to build a detailed profile. It begins by categorising them into a general Player Archetype (e.g., Rock, Fish, Calling Station) based on their core statistics. This classification is based on plotting the opponent on a two-dimensional matrix of playing style, defined by their tendencies for looseness (VPiP) and aggression (Aggression Factor), as illustrated in Figure 6.

In the Destroy phase, the module weaponizes this intelligence. It formulates a specific, high-conviction exploit and recommends it to the Master Algorithm. For example, if it discovers a player folds to continuation bets on the turn 80% of the time, it will send a high-conviction recommendation to 'double barrel' bluff against that player, regardless of Patrick's own hand strength.

This module is what elevates Patrick from a fundamentally sound player into a truly dangerous and adaptive opponent, capable of systematically dismantling an opponent's strategy piece by piece.

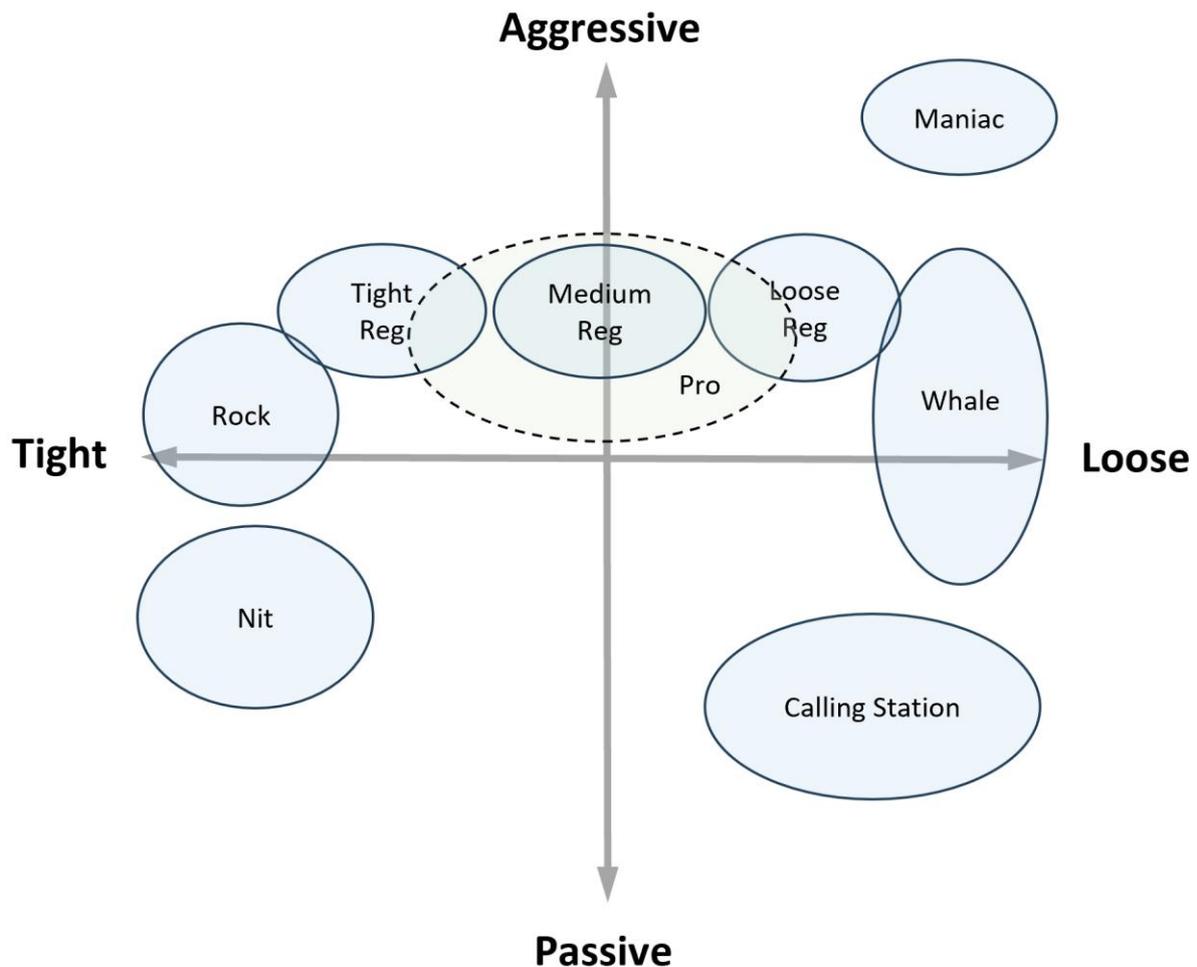

Figure 6: A conceptual model of player archetypes, plotted by VPiP (looseness) and Aggression Factor (aggression).

***The Lawnmower***

The Lawnmower is the module responsible for executing the system's most sophisticated, multi-layered deceptive plays. While the Ranges module focuses on Level 1 thinking (deducing an opponent's holdings[4]), The Lawnmower operates at a higher level of abstraction: Level 2 thinking. It models how an opponent is likely to perceive *Patrick's* actions.

By simulating the opponent's own hand-reading process—a simulation informed by the Player Archetype assigned by the Search and Destroy module—it can craft lines of play designed to tell a specific, misleading story. This is the module that directs the sophisticated human emulation behaviours, such as orchestrating complex, multi-street bluffs or using 'Hollywood' pauses to feign a difficult decision when holding the nuts. It is the component that best encapsulates the project's exploitative philosophy, turning poker from a mathematical exercise into a game of psychological warfare.

***The Master Algorithm (MA)***

The Master Algorithm (MA) is the final arbiter—the 'conscious mind' of the system. It receives a confluence of suggestions from the various decision modules, each with its own perspective and conviction. Its role is not simply to resolve consensus, but to synthesise this information, expertly weighing the options it is presented with in the context of the current game state.

For example, in a hand where Patrick holds the nuts against multiple opponents, the General Algorithm, based on the standard principle of deception, might suggest a 'Call' to encourage other players to stay in the pot. In parallel, 'Search and Destroy' would analyse the opponent who made the initial bet, identify them as a 'Whale', and based on historical data, determine that this player type is highly likely to call a raise with any marginal hand. It would therefore send a high-conviction recommendation to 'Raise', with the specific strategic goal of pot-committing this weak opponent early, ensuring their entire stack is at risk for the remainder of the hand.

The Master Algorithm now has conflicting advice: a deceptive 'Call' versus an aggressive 'Raise'. Its role is to weigh these inputs, and in a situation where a specific, profitable weakness has been identified, it gives a higher priority to the targeted, exploitative recommendation from 'Search and Destroy'. Therefore, it makes the final decision to raise.

As a core principle, the Master Algorithm ignores past financial results, as they are rendered meaningless by variance. It can also instigate overarching strategies for an entire hand, such as adjusting the tempo of its play based on factors like the effective stack size. Ultimately, it is the component that calmly assesses all strategic inputs to make the final, decisive action.

## Case Study: The Ranges Module in Action

Having detailed the theoretical architecture of the Brain, the best way to understand its practical application is to follow a single, challenging example from the trial (Hand 6). This case study will focus on the Ranges module, demonstrating the *process* of range-narrowing in real time to show how an initial wide distribution of possibilities is refined into a precise final read.

This hand was chosen specifically because it represents a formidable challenge for hand-reading: an unorthodox action from an unpredictable Player Archetype (a Whale), creating a situation where even elite human players would struggle to get an accurate read. The process mirrors the logic of an expert human player: assign a range of possible hands, then narrow those possibilities as each action provides new information.

### The Theory: What an Action Means

Before analysing a hand, it is crucial to understand the theory of what a player's action represents. For a GTO *solver*, an action is a singular, mathematical decision. For a human, an action is the result of a complex and often messy internal monologue. Patrick's architecture is designed to navigate this human reality, understanding that a single action can stem from wildly different motivations, ranging from the highly strategic to the purely emotional.

A 'bet', for example, can be driven by entirely different thought processes. At a strategic level, it could be a logical bet for value, where the monologue is 'I believe I have the best hand and I want to be called'. Conversely, it might be a calculated, strategic bluff, where the thinking is 'I believe my opponent has a better hand, but I can tell a convincing story that will make them fold'. At the other end of the spectrum, however, the motivation can be a purely irrational, emotional bluff, where the monologue is simply 'I will not be bullied again'. This is not a strategic play but an emotional reaction, often driven by frustration, ego, a misunderstanding of the game state, or what poker players refer to as *tilt*.

Similarly, a 'check' can be equally ambiguous, signalling anything from genuine weakness ('I give up'), to a form of pot control with a medium-strength hand, or even a deceptive trap with a monster holding. The 'call' is perhaps the most context-dependent action of all, representing a vast spectrum of skill and intent. For an advanced player, it might be a sophisticated manoeuvre known as 'floating' (calling in position to set up a future bluff). For a competent player, it is often a standard, logical play based on pot odds to chase a draw. For a novice, however, a call is frequently driven by flawed logic: a stubborn refusal to be bullied, a mistaken sense of being 'pot committed', or simple, undisciplined curiosity to 'see what they have'.

Navigating this ambiguity is the central challenge Patrick's architecture is designed to solve. Where a GTO-based system might interpret an unorthodox action as a simple mathematical deviation, Patrick's modules are built to diagnose the underlying human cause. The Search and Destroy module, for instance, builds a detailed statistical profile of an opponent to determine whether their actions are more likely to be driven by sound strategy or by predictable, exploitable tendencies. This allows the system to move beyond a generic response and formulate a specific counter-strategy targeted at the individual's psychological profile.

To see this theory in practice, consider a clear, illustrative example: a strong, last-to-speak bet from a *competent* opponent on the river. This single action dramatically polarises the opponent's likely holdings. As the poker adage goes, 'he's either got 'em or he ain't'. This means the opponent's range is concentrated at two extremes: very strong hands played for value, or weak hands played as a bluff. It excludes medium-strength hands, for which the likeliest action would be to check for a showdown.

This logic is captured in what the system calls a **Range Reshaping Template (RET)**. An RET is a predefined probabilistic model that is applied to an opponent's range following a specific action. For any given action, it assigns a unique likelihood weighting to each category of hand strength—from Niente to the Nuts—thereby reshaping the probabilities of the hands an opponent might hold.

For instance, the specific RET for the strong, last-to-speak river bet is heavily polarised; it increases the weighting of the polar extremes (very strong hands and bluffs) while heavily discounting the middle, as shown in the theoretical model below.

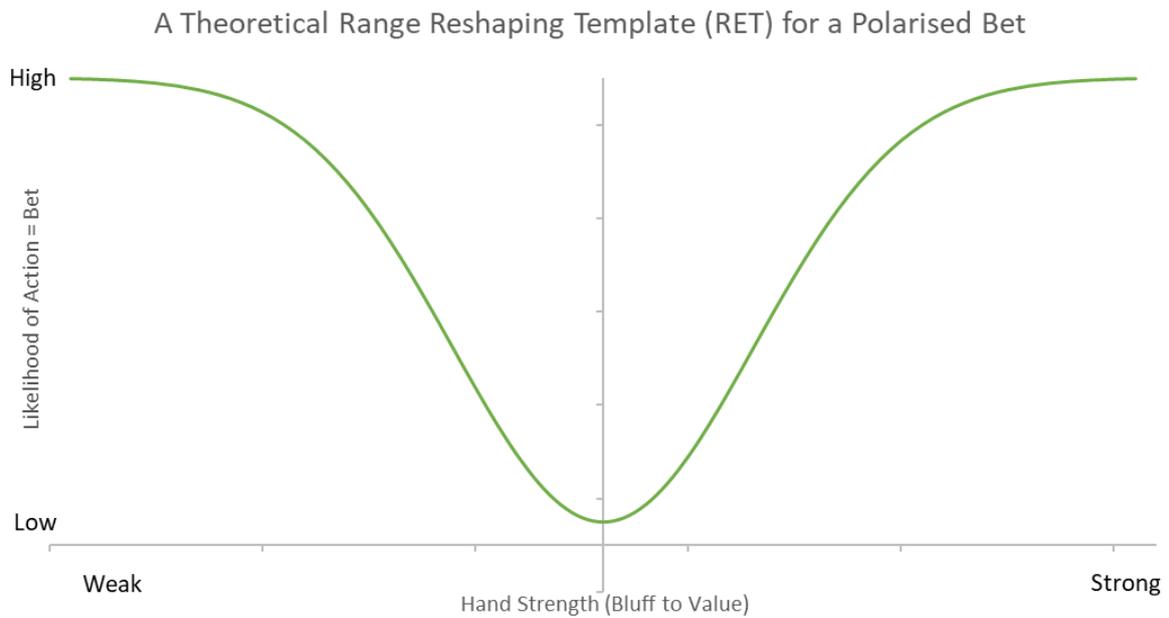

Figure 7: A theoretical Range Reshaping Template (RET) for a strong bet, showing a high likelihood of very weak or very strong hands, and a low likelihood of middling hands

**The Relative Strength (rS) Distribution**

After every action or new community card, Patrick analyses an opponent's likely holdings through a model known as the Relative Strength (rS) distribution. This distribution is a probabilistic snapshot of an opponent's entire hand range at a given moment.

It works by categorising every possible hand in their range onto a simple, 11-point scale of strength, from 0 (**Niente** – *nothing*), through the mid-points like **Fair** and **Good**, up to 9 (**Nuts**) and 10 (**Alcatraz**). The term *Alcatraz* is reserved for hands like four of a kind which, while unbeatable, cripple the board and thus significantly reduce the likelihood of action.

The resulting bar chart shows the percentage chance that an opponent's hand falls into one of these categories. The primary purpose of this distribution is to calculate the Chance I'm Beat (ChiB), a crucial metric that is passed to the Master Algorithm to inform its final decision.

**The Application: A Step-by-Step Analysis of Hand 6**

*Pre-flop*

The hand begins with Patrick holding the 9♥9♠ and making a standard raise from Under the Gun (UTG). A player identified as a Medium-Regular calls from the small blind, and a Whale calls from the big blind. This analysis will focus on the Whale, as his unorthodox play provides the more instructive demonstration of the Ranges module. Based on his Player Archetype and the excellent pot odds offered, the module assigns a very wide initial starting range, as shown in Figure 8.

This 13x13 grid represents all 169 possible starting hands, with pocket pairs on the diagonal, suited hands above, and offsuit hands below. The colour of each square acts as a heatmap, where a darker shade indicates a higher probability. This shading accounts for the underlying combinatorics; for instance, there are twelve combinations of an offsuit hand like QJo versus only four for its suited counterpart, QJs, making it three times more likely to be held.

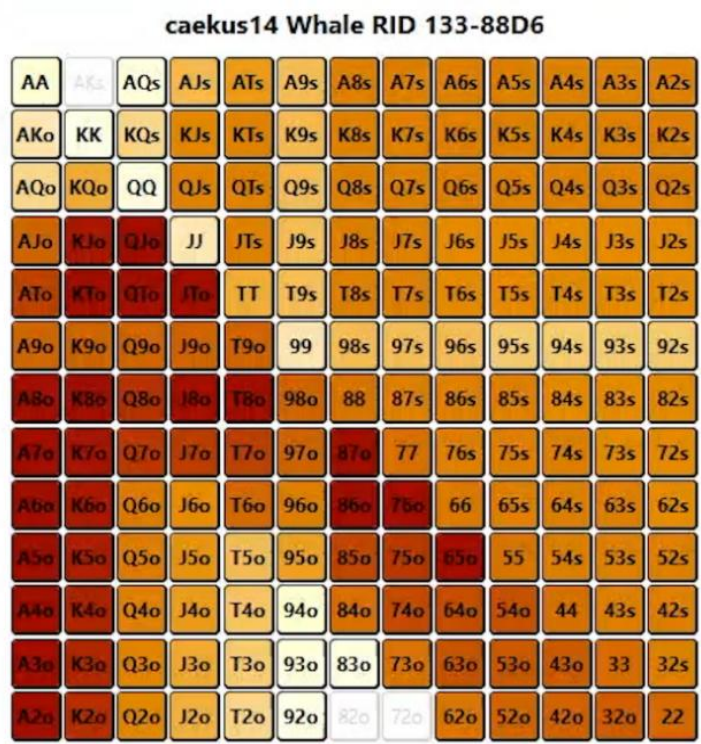

Figure 8: The Whale's pre-flop range. *Heatmap: Darker shading indicates a higher probability.*

*Flop*

The flop is 9♦5♣2♣, giving Patrick top set. The system now strips these community cards from the opponents' possible holdings, which updates the Whale's range grid as shown in Figure 9.

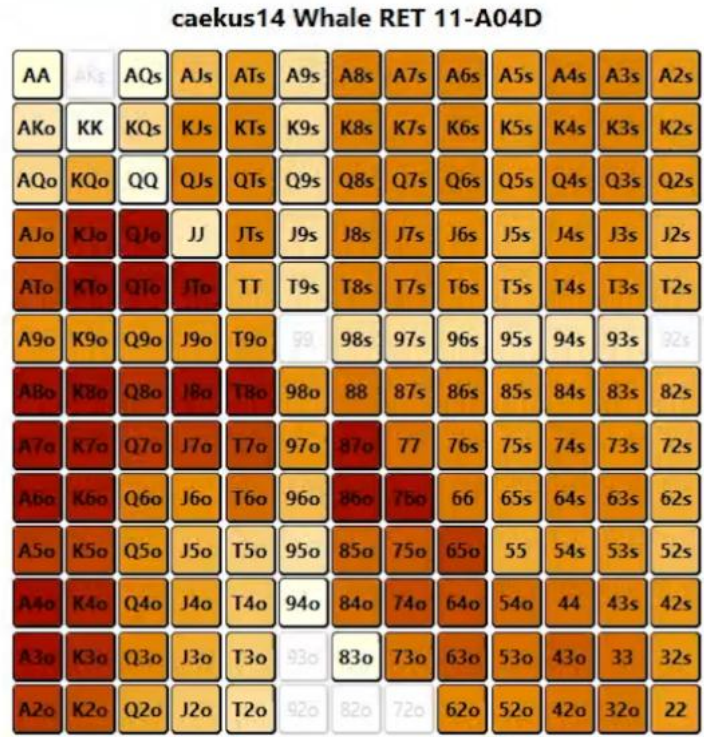

Figure 9: The Whale's hand range after the flop cards are stripped. *Heatmap: Darker shading indicates a higher probability.*

Each hand combination within this updated range is then processed by the Relative Strengths Matrix (RSM) to generate the initial post-flop rS distribution, shown in Figure 10. As expected, this distribution reveals that the Whale's wide pre-flop range has connected poorly with the board; it is heavily weighted toward weak holdings. Niente and HardlyAnything collectively account for 61% of his probable hands, with only a 12% chance of holding a hand of Fair strength or better.

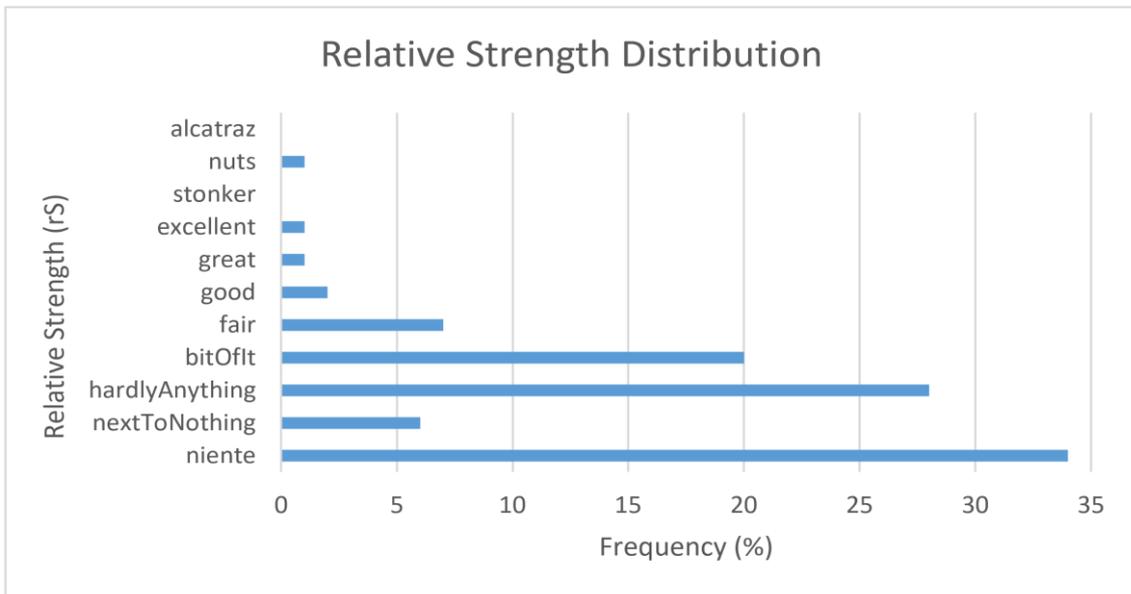

Figure 10: The Whale's initial rS distribution on the flop
(processed with a flat RET).

This initial analysis provides an excellent opportunity to contrast the strategic implications of the two different playing styles at the table. In stark contrast to the Whale, the Medium-Regular in the small blind, who started with a much narrower range, is in a much stronger position (Figures 11 & 12). His rS distribution shows a 0% chance of holding Niente and a 7% chance of holding the Nuts, more than three times as likely as the Whale. This highlights a fundamental trade-off: the Regular's narrow range connects more strongly with the board but makes him more predictable; the Whale's wide range connects far less often but makes him much harder to read.

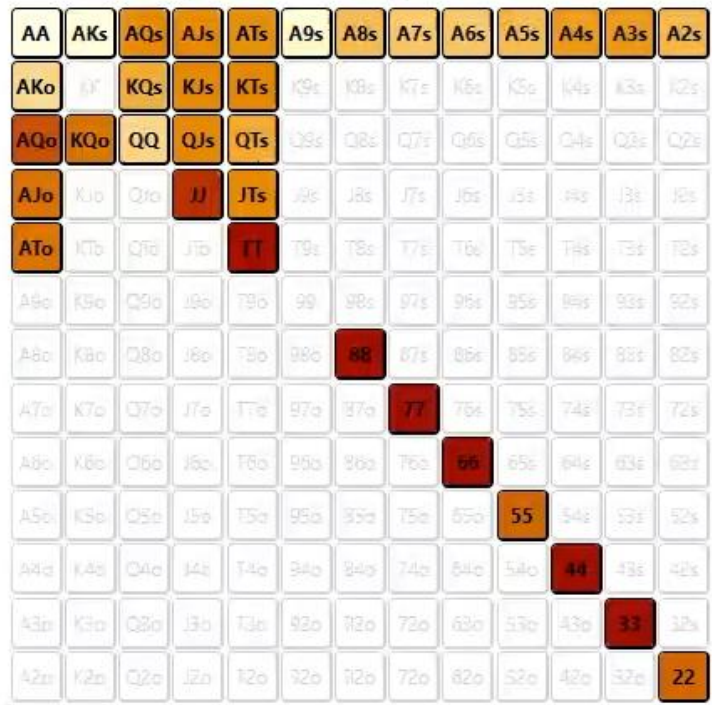

Figure 11: The Medium-Regular's narrow hand range after the flop cards are stripped. *Heatmap: Darker shading indicates a higher probability.*

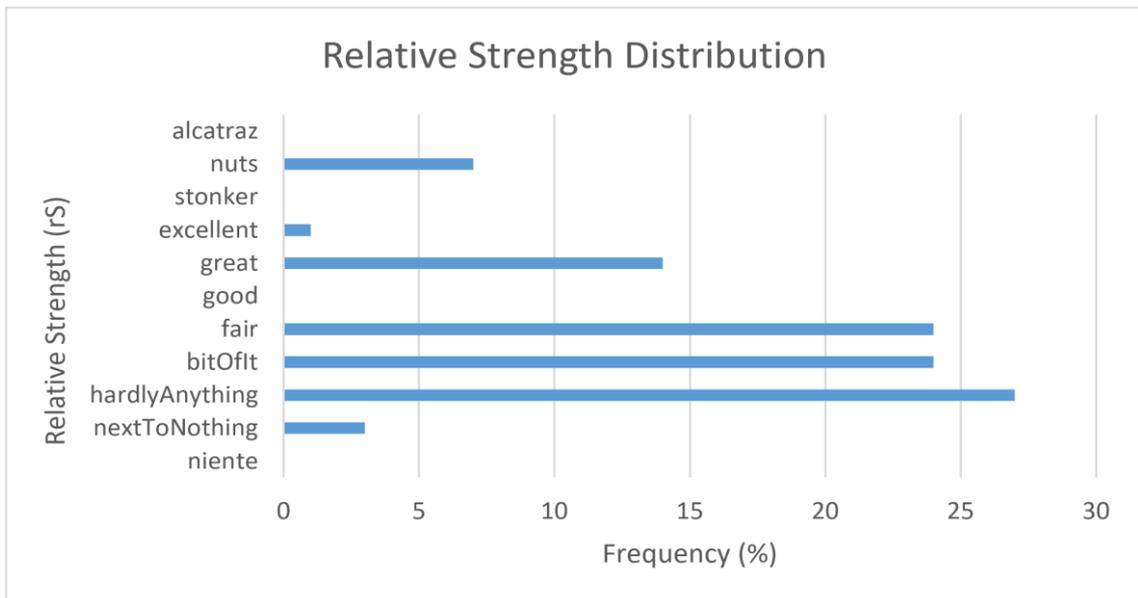

Figure 12: The Medium-Regular's initial rS distribution on the flop (processed with a flat RET).

*Flop Action*

The Small Blind checks, and the Whale makes an unorthodox 'donk bet'. This action is where the theoretical model of range reshaping is put into practice. The system processes the bet not with a generalised curve, but with a specific, granular template: RET 18 (Figure 13).

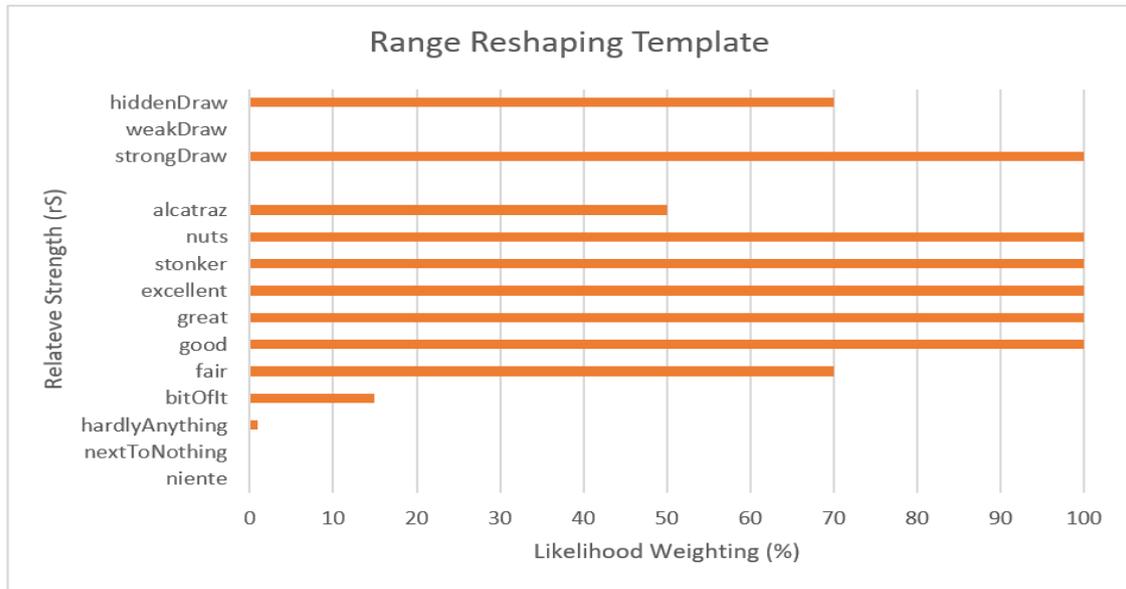

Figure 13: RET 18: The reshaping template for the
Whale's donk bet on the flop

As the chart demonstrates, this template is heavily weighted toward strong made hands and key draws. For example, it assigns a high likelihood to Excellent hands like big overpairs, Great hands like a medium overpair or top pair with a strong kicker, and Nuts such as a set of fives. It also strongly weights valuable draws, such as the open-ended straight draw held by a hand like 43. Applying this template produces the updated range grid (Figure 14) and rS distribution (Figure 15).

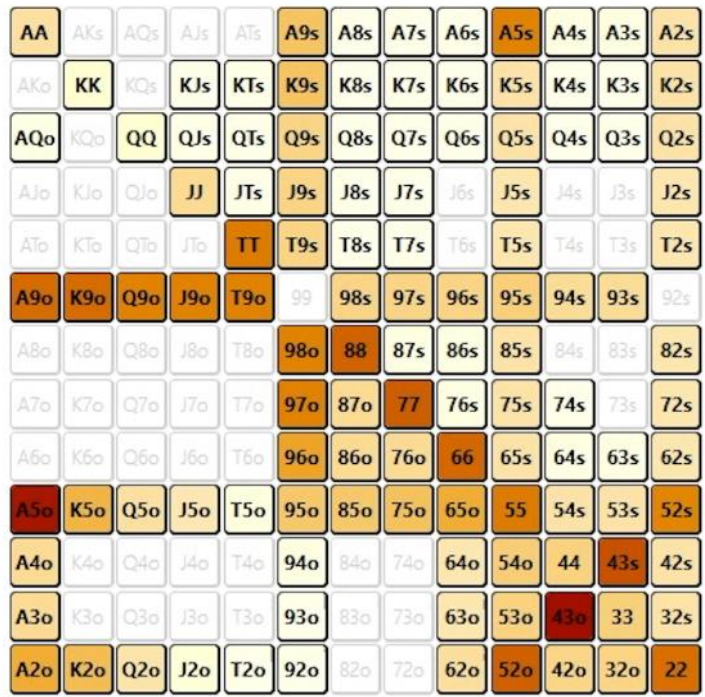

Figure 14: The Whale's range, reshaped by his decision to bet. *Heatmap: Darker shading indicates a higher probability.*

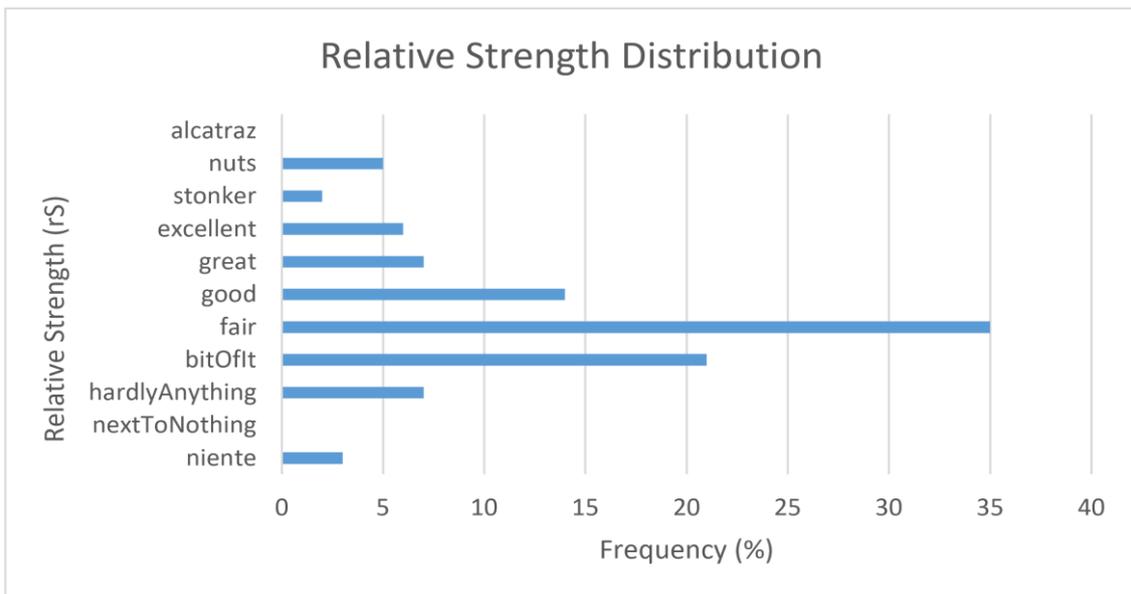

Figure 15: The Whale's rS distribution after his donk bet on the flop (RET 18).

Patrick then raises, and the Whale calls. This subsequent action is processed with RET 33 (Figure 16). The resulting analysis provides a crucial insight: having now declined two opportunities to re-raise (pre-flop and on the flop), it is much less likely that the Whale holds one of the strongest possible hands. The range is now reshaped to favour medium-strength made hands and strong draws that would logically just call in this situation. The final range grid for the flop (Figure 17) and the final rS distribution (Figure 18) reflect this new assessment, and the updated ChiB is passed to the Master Algorithm.

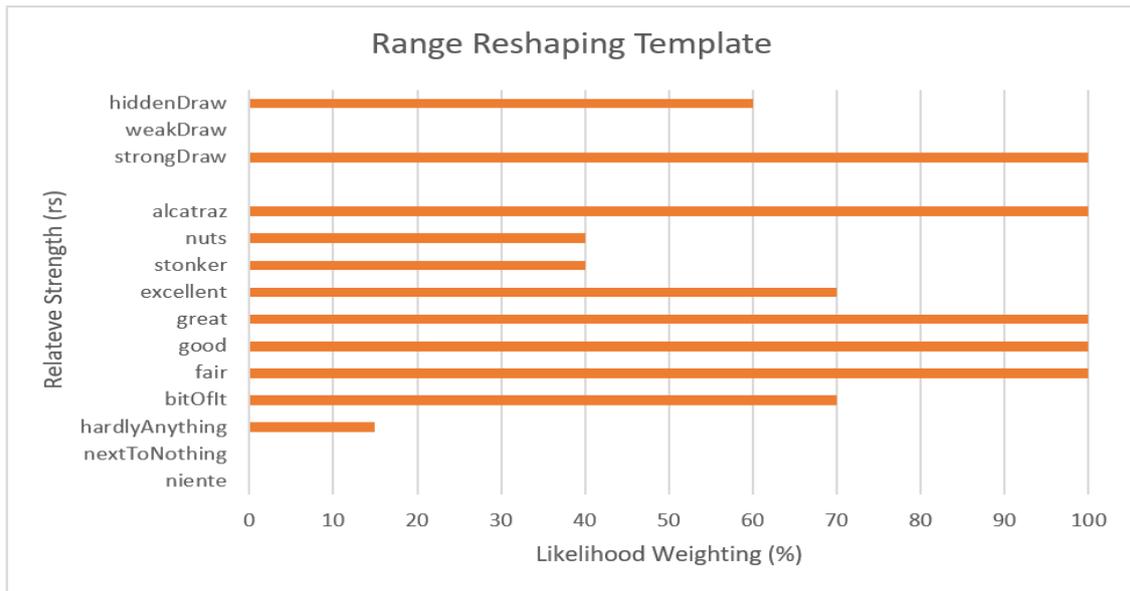

Figure 16: RET 33: The reshaping template for the Whale's call on the flop.

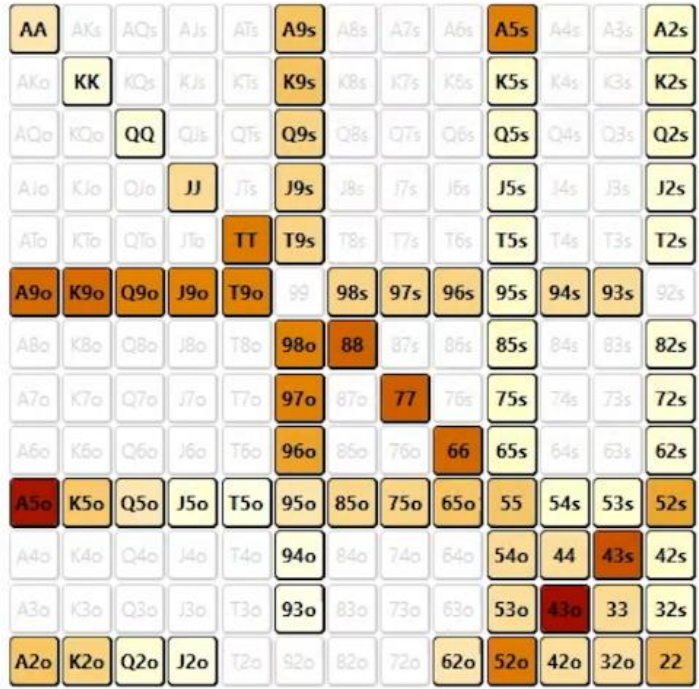

Figure 17: The Whale's final flop range after calling Patrick's raise. *Heatmap: Darker shading indicates a higher probability.*

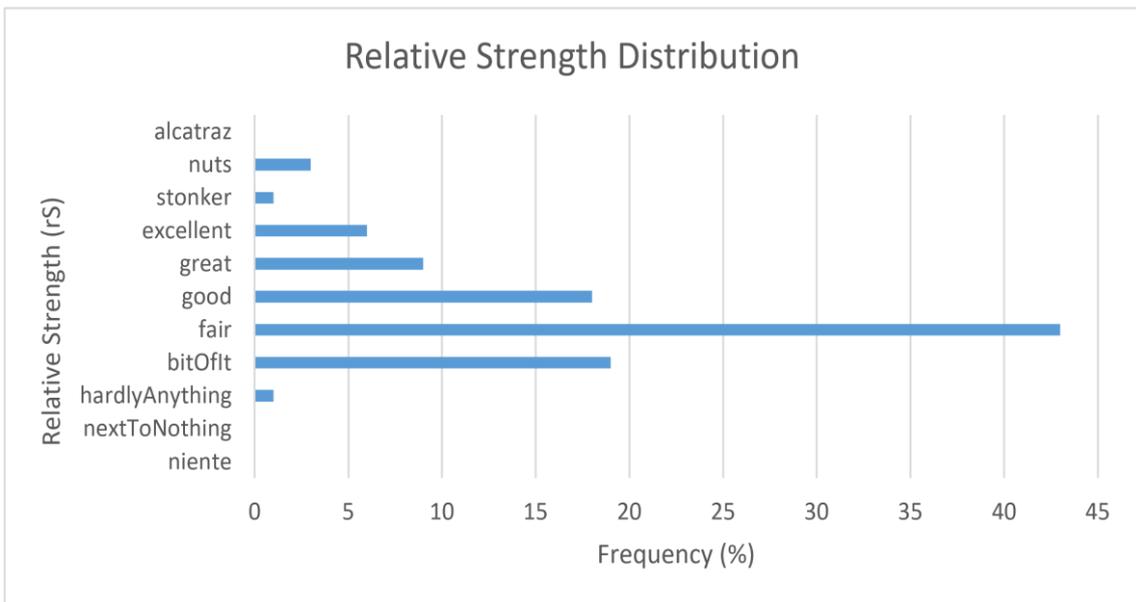

Figure 18: The Whale's rS distribution after calling the raise on the flop (RET 33).

*Turn*

The turn is the 2♦, pairing the board and completing Patrick's full house. The system first updates all ranges to account for this new community card (a process using a flat RET), producing a new rS distribution and an updated ChiB.

The Whale then bets all-in. This final action is analysed with its own specific template, RET 73, to produce the final read before Patrick makes the call.

*Showdown*

As the players are all-in, the cards are revealed. The river card, the K♠, is dealt but is irrelevant to the outcome as Patrick already holds the winning hand. The Whale shows 4♦3♦ for a busted straight and flush draw, confirming the accuracy of the system's final analysis.

Board 9♦ 5♠ 2♣ 2♦ K♠

| Player  | Hand   |
|---------|--------|
| Patrick | 9♥ 9♠ |
| Whale   | 4♦ 3♦ |

Table 3: Showdown results

*Strategic Analysis*

This hand highlights a core principle of the system's methodology: the accuracy of its read increases with each subsequent action. The initial pre-flop range assigned to the Whale is very wide and uncertain, but with each check, bet, or call, the range is refined and narrowed, leading to a much more accurate and confident final assessment.

Furthermore, Patrick's large raise on the flop was a forward-thinking, strategic play designed to induce an error. By making the pot large early, it successfully pot-committed the Whale, who held a strong combination draw. Because the Whale was already committed, his best option on the turn was a semi-bluff shove, even on a dangerous paired board. Patrick's initial raise created this exact situation, serving as a clear demonstration of an aggressive, multi-street strategy designed to exploit an opponent's tendencies.

A granular, card-by-card breakdown of this hand, including the most detailed range visualisations, is available in Appendix C.

## Machine Learning Methodology

The previous section concluded with a fundamental question: how can an AI learn to master the art of human imperfection without a reliable feedback mechanism? Patrick's machine learning methodology is the answer. It is a dual-process system designed to extract maximum value from the limited information available in a real-world environment, adapting to the reality that opponents' cards are rarely revealed.

**Learning Without Showdowns: Continuous Statistical Profiling**

In the vast majority of hands, opponents' cards are not seen. In these cases, learning is focused on continuous statistical profiling. For every action an opponent takes, Patrick updates a detailed profile, tracking dozens of statistical tendencies such as pre-flop raise frequency (PFR) and continuation betting frequency. This method, analogous to commercial poker tracking software, is the primary intelligence-gathering mechanism for the Search and Destroy module. It allows the AI to build a rich, data-driven model of each player's style over time, which is then used to identify and target specific vulnerabilities.

**Learning From Showdowns: Deep Model Refinement**

When cards are revealed at a showdown, a deeper, model-level learning process is triggered. The system first checks if the revealed hand was within the range it had assigned to that opponent. If not, it substantially adjusts its range model for that specific player and also applies a minuscule adjustment to its model for all players of that archetype. This initial check ensures the AI's opponent modelling becomes more accurate over time.

**A Novel Approach: Anchoring on Predictions, Not Results**

Crucially, Patrick's machine learning model deliberately ignores the financial result of a hand. A positive monetary outcome is not necessarily indicative of good play, and a negative one does not imply poor play; a machine cannot distinguish a well-earned pot from a lucky win without a human-level understanding of context.

Instead of anchoring on unreliable results, Patrick anchors on the accuracy of its own internal predictions. After a showdown, the AI re-runs the hand with perfect information, comparing its predictions at each decision point with the ground truth. This feedback loop updates core architectural components, most notably the RSM. A small reinforcement delta is applied where predictions were correct, and a larger corrective delta is applied where they were wrong. This process, based on predictive accuracy rather than monetary results, allows the AI to improve its decision-making framework without being misled by short-term variance. This approach draws inspiration from reinforcement learning techniques but adapts them for the specific challenges of an incomplete information environment.

# Conclusion

This paper began by challenging the orthodoxy of defensive, *solver*-based AI, proposing an alternative philosophy: that the path to victory in a human-centric game lies not in being unexploitable, but in being maximally exploitative. The profitable outcome of the 64,267-hand trial serves as a validation of this 'sword'-based approach. Patrick's success is not a declaration that poker is *solved*, but rather a testament to an architecture designed to master the art of human imperfection.

The project also concludes with a deeper humility for the obstacles involved. The profound effects of variance, so often rooted in the chaos of human emotion, teach us that short-term results are an illusion, while the constant challenge of system stability reminds us of the gap between a clean algorithm and a chaotic real-world environment.

The development of an AI that learns from its predictions, not its winnings, and is built to target psychological flaws, represents an important step forward. More importantly, it offers a new perspective on the challenges and opportunities that lie on the path toward creating truly intelligent, adaptive systems that can thrive not in a sterile laboratory, but in the complex and unpredictable world of human interaction, suggesting the next frontier is not about creating perfect shields, but about forging ever-sharper swords.

## Limitations

It is important to contextualise the results of this trial within the scope of its limitations. The trial was conducted exclusively in the 1¢/2¢ micro-stakes environment. While this player pool is highly varied and unpredictable, it is not representative of the more strategically uniform and technically proficient opponents found at higher stakes.

Furthermore, while the sample of over 64,000 hands is sufficient to demonstrate a statistically significant win rate, poker is a game of high variance. A larger sample, potentially 200,000 hands or more, would be required to achieve a higher degree of confidence in the long-term sustainability of this win rate. These factors present clear areas for future research

## Future work

The success of Patrick serves not as a conclusion, but as a proof of concept for a more ambitious, long-term research agenda. Patrick's architecture was purpose-built to model and exploit the strategic vulnerabilities of human opponents within the closed system of poker. Future work will seek to generalise this 'sword-based' philosophy beyond a single game, focusing on the development of a framework for modelling the complex heuristics that underpin human decision-making in broader, less-structured environments.

This next-generation architecture will move beyond identifying purely strategic patterns to explore the cognitive and, ultimately, the emotional drivers of human behaviour. The central research question will shift from 'What is the opponent's most likely action?' to 'What is the underlying cognitive state that motivates that action?'

The ultimate aim is to create systems that possess a more nuanced and predictive understanding of human interaction. Such an AI could offer significant benefits in fields requiring complex human-machine collaboration, from advanced decision-support systems to more intuitive and adaptive training tools. This continued study of human imperfection suggests it is possible to create AI that can not only master a game, but can also become a more effective partner in navigating the challenges of the real world.

## Acknowledgments

The authors would like to thank independent reviewers for their valuable feedback on earlier drafts of this paper.

# Appendix A: Detailed Financial Results

Table A.1: Detailed Financial Results

| Game | 1c-2c 6-handed fast fold Holdem - 888 Poker.com | Hands Played |
|---|---|---|
| Trial period | 1st January 2023 to 26th February 2023 | 64,267 |

| Results | Amount won (excluding rake and rakeback) | Rake Attributed | Rakeback | Amount Won (including rake, excluding rakeback) | All-in Adjusted (excluding rakeback) | Bank Balance (true amount won, including rake and rakeback) |
|---|---|---|---|---|---|---|
| Cash | $177.92 | -$138.70 | $9.59 | $38.22 | $56.18 | $47.81 |
| BB/100 | 13.8 | -10.8 | 0.7 | 3.0 | 4.4 | 3.7 |
|  | The measure often used in scientific experiments (Liberatus and Pluribus played without rake and rakeback) | Average rake attributed for the field during period: 13.0 BB/100 |  | Pokertracker's prefered measure | The measure sometimes used by professional poker players | The true amount won |

## Appendix B: Glossary of Terms

The following is an alphabetical list of key terms used throughout this paper. Terms developed as part of Patrick's architecture are **bolded**.

Absolute Strength: The ranking of a five-card poker hand according to the standard hierarchy (e.g., a full house, a flush, a straight). This value is static and does not change based on context. It is one of the two primary inputs for the **Relative Strengths Matrix (RSM)**.

Abstraction Layer: In software engineering, an abstraction layer is a mechanism that hides the complex, low-level workings of a system behind a simplified, high-level interface. Within **Patrick's** architecture, the **RSM** acts as an abstraction layer by translating the immense complexity of board textures and absolute hand strengths into a single, high-level concept (a hand's relative strength), which can then be easily used by other modules like the **GA**.

Aggression Factor (AF): A key statistic provided by poker tracking software that measures a player's post-flop aggression. It is calculated as the ratio of their aggressive actions (bets and raises) to their passive actions (calls). A high Aggression Factor indicates a player who frequently bets or raises rather than just calling. Along with **VPiP**, it is a primary metric used to determine a **Player Archetype**.

AIVAT (All-in Adjusted Value): A metric used in some academic studies to measure an AI's performance based on its strategic choices, adjusting for the luck of all-in situations. This paper posits that this metric can be misleading when evaluating an exploitative strategy.

All-in Adjusted: A metric displayed by poker tracking software that calculates a player's expected winnings based on their equity at the moment they went all-in, rather than the actual result. It is often referred to as the "luck line."

Bad Beat: A term for losing a hand in which the player was a strong statistical favourite to win.

BB/100: "Big Blinds per 100 hands." The standard metric for measuring win rate in online poker. A Big Blind is the larger of the two forced bets that initiate a hand.

Blinds: Forced bets posted by two players before the cards are dealt (the Small Blind and the Big Blind) to ensure there is money in the pot to play for.

Big Blind: The larger of the two forced bets posted before the cards are dealt. It is the fundamental unit of measurement for bet sizes and win rates (see **BB/100**).

Blank: A community card, typically on the **Turn** or **River**, that is unlikely to have improved any player's hand or completed any obvious draws.

Board Texture: A qualitative term describing the characteristics of the community cards on the table. A "draw-heavy" or "wet" board (e.g., J♠ T♠ 7♣) contains many potential straight and flush draws, making it dangerous. A "dry" or "uncoordinated" board (e.g., K♦ 8♣ 2♥) offers few draws, making strong made hands more valuable. It is one of the two primary inputs for the **Relative Strengths Matrix (RSM)**.

Button: The position at the poker table that acts last in the post-flop betting rounds, giving them a significant strategic advantage.

Calling Station: A specific type of **Fish** characterised by a highly passive style. A Calling Station will frequently call bets with a wide range of hands but will rarely raise or fold, making them a primary target for repeated value bets.

**Chance I'm Beat (ChiB):** A crucial metric calculated by the Ranges module that represents the real-time probability of **Patrick**'s hand being weaker than an opponent's estimated range.

Complete Information: In game theory, a game of Complete Information is one in which all players have perfect knowledge of the game state at all times. Chess is a canonical example, as there is no hidden information; all pieces are visible to both players. Within the context of this paper, the term is also used to describe the condition created in a laboratory setting where an analyst or learning algorithm is granted access to all hidden information from a hand *after* it has been played. This post-hoc creation of a complete information state is a luxury unavailable in real-world play and is a prerequisite for certain evaluation metrics, such as **AIVAT**

Continuation Bet (C-Bet): A bet made on the **Flop** by the player who was the pre-flop aggressor (the last person to bet or raise before the flop).

Counter-factual Regret Minimisation (CFR): An iterative algorithm used by AIs like *Libratus* to find optimal strategies in games of imperfect information by minimising potential "regret" over past decisions.

Donk Bet: An unorthodox bet made by a player who is out of position and was not the aggressor on the previous street.

Double Barrel: The act of making a second consecutive bet on the **Turn** after having already made a **Continuation Bet** on the **Flop**. The term can be extended to "triple barrel" for a third bet on the **River**.

Equity: A player's statistical share of the current pot, calculated based on their probability of winning the hand at a given point. For example, if a player has an 80% chance of winning a $100 pot, their equity in that pot is $80. This metric is the basis for the **All-in Adjusted** calculation.

Fast-Fold Poker: An online poker format where players are immediately moved to a new table with a new hand as soon as they fold, allowing for a much higher volume of hands per hour.

Fish: A general term for an unskilled, highly exploitable player, typically characterised by a high **VPiP** (playing too many hands). It is a broad player archetype that includes more specific sub-types, such as the **Calling Station**.

Flop: The first and most defining stage of post-flop play, in which the first three community cards are dealt, followed by a round of betting.

Game Theory Optimal (GTO): A style of play, often pursued by *solvers*, that aims to be mathematically unexploitable. **Patrick**'s "sword" philosophy is presented as an alternative to this defensive "shield" approach.

**General Algorithm (GA):** The baseline strategic engine of the Brain; an expert system that plays fundamentally sound poker based on core mathematical principles and heuristics.

Green Line: A universally understood convention for the line on a Poker Tracker graph that displays the net profit or loss from hands played at the table. This metric is the standard measure of on-table performance, as it includes the negative impact of **rake** paid but excludes external factors such as **rakeback**. It is therefore distinct from the final change in a player's bank balance and is visually distinguished from the **yellow line**.

**Hand Approach Algorithm (HAA):** A supporting module in the Brain that assesses the game state at the start of each hand to make subtle adjustments to **Patrick**'s baseline strategic style, including simulating states like 'tilt' for camouflage.

Hand Notation: The standard shorthand used to describe a player's two-card starting hand. Card ranks are represented by A (Ace), K (King), Q (Queen), J (Jack), T (Ten), and numerals (9-2).

- Suited Hands: An 's' appended to two card ranks indicates they are of the same suit (e.g., 23s is a two and a three of the same suit; AKs is an Ace and King of the same suit).

- Offsuit Hands: An 'o' appended to two card ranks indicates they are of different suits (e.g., ATo is an Ace and Ten of different suits).

- Pocket Pairs: Two identical card ranks represent a pair (e.g., 99 is a pair of nines; AA is a pair of Aces).

Heads-Up: A poker hand or game that is contested by only two players.

Incomplete Information: A game in which players do not have full knowledge of the game state. Poker is the canonical example, as players' private 'hole cards' are hidden from their opponents. This lack of information forces players to make decisions based on probability, psychology, and deduction, which is the central challenge the **Patrick** AI was designed to navigate

Kicker: An unpaired card in a player's hand that does not contribute to a pair or other combination but can be used to break ties (e.g., holding Ace-King on a King-high board gives you "top pair, top kicker").

**KuKulKan Grid:** A highly granular 1,326-square grid that visualises an opponent's hand range by representing every possible two-card starting hand combination.

**Lawnmower:** The decision module responsible for executing sophisticated, multi-layered deceptive plays by modelling how an opponent is likely to perceive Patrick's actions (Level 2 thinking).

Loose / Tight: Adjectives describing a player's tendency to play a wide (loose) or narrow (tight) range of starting hands, often measured by **VPiP**.

Level 1 Thinking: A level of strategic thought focused solely on the strength of one's own hand (e.g., "I have a strong hand, so I will bet"). The **Ranges** module's function of deducing an opponent's hand is an advanced form of this.

Level 2 Thinking: A more advanced level of strategic thought focused on what an opponent is likely to think about *your* hand based on your actions (e.g., "If I check now, my opponent will think I am weak and will try to bluff me"). This is the primary operational domain of **The Lawnmower** module.

**Loose Aggressive (LAG):** A standard player archetype characterised by a high **VPiP** (playing a wide range of starting hands) and an aggressive post-flop style.

**Master Algorithm (MA):** The final arbiter of the Brain, responsible for synthesising recommendations from all decision modules to make the final, decisive action.

Medium Aggressive (MAG): A standard player archetype whose style falls between a **Tight-Aggressive (TAG)** and a **Loose Aggressive (LAG)** player, characterised by playing a moderate range of hands but doing so aggressively.

Micro-Stakes: The lowest levels of online poker, typically with blinds of $0.01/$0.02 or $0.02/$0.05, which feature a highly diverse and unpredictable player pool.

Nash Equilibrium: A foundational concept in game theory where no player can benefit by changing their strategy while the other players keep their strategies unchanged. It is the theoretical goal of GTO play.

**Niente:** The lowest possible rating (0) on the Relative Strength scale, indicating a hand with no value (e.g. 8-high on the **river**).

Overpair: A pocket pair that is higher in rank than any of the cards on the flop.

**Patrick:** The adaptive poker AI developed by Spiderdime Systems, built on the philosophy of being maximally exploitative rather than mathematically unexploitable.

PFR (Pre-Flop Raise): A key statistic that measures how frequently a player makes a raise before the flop when they have the opportunity. Paired with VPiP, it helps define a player's aggression.

Player Archetype**:** A classification of a player based on their statistical tendencies (e.g., **Rock**, **Calling Station**, **Regular**). **Patrick**'s **Search and Destroy** module categorises opponents into these archetypes for targeted exploitation.

Poker Tracker: A cornerstone of the modern online poker toolkit, Poker Tracker is a sophisticated database application that imports and parses hand histories. Its function is twofold. Firstly, it enables rigorous post-session analysis, allowing players to identify strategic patterns and systematically improve their own game. Secondly, and more critically for real-time play, it provides a **Heads-Up Display (HUD)**, which overlays detailed statistical profiles directly onto the table. This provides the crucial data that enables a human player to move beyond general strategy and execute a targeted, exploitative style against the specific, observable tendencies of each opponent. It is widely considered an indispensable tool for serious students of the game and professional players.

Position: A player's location relative to the dealer button. Acting after an opponent ("in position") is a major strategic advantage.

Pot Odds: The ratio of the current size of the pot to the cost of a contemplated call. This ratio is a fundamental part of poker mathematics, used to determine if calling with a drawing hand is profitable.

Pre-flop Aggressor: The player who made the last bet or raise during the pre-flop betting round. This player is said to have the "initiative" going into the post-flop streets and is often expected to make a **Continuation Bet** on the **Flop**.

Rake: The commission fee taken by the poker site from each pot played. It is the primary cost of playing and has a profound impact on profitability.

Rakeback: A loyalty incentive offered by poker sites where a small percentage of the rake a player has paid is returned to them.

**Range Reshaping Template (RET):** A predefined probabilistic model that is applied to an opponent's hand range after they take an action. Each RET assigns likelihood weightings to different hand strengths, thereby narrowing the possibilities of what the opponent holds.

Regular (Reg): A player archetype for a competent and experienced player who plays a solid, predictable strategy. They are often further classified by their playing style (e.g., Tight Reg, Medium Reg, Loose Reg) and are a common fixture in the player pool.

**Relative Strength (rS):** A precise value (from 0 to 10) that represents the strength of a given hand in the context of the specific board texture and game situation, as opposed to its absolute value.

**Relative Strengths Matrix (RSM):** A multi-dimensional, self-correcting data model that outputs the Relative Strength value for any given hand and board combination. It is the foundational data model of the Brain.

River: The final stage of post-flop play, in which the fifth and last community card is dealt, followed by the final round of betting. It is also known as 'fifth street'.

Rock: A player archetype characterised by an extremely tight style, playing very few hands and typically only entering the pot with premium holdings.

**Search and Destroy (SAD):** The decision module responsible for targeted exploitation. It operates in a two-phase cycle, first searching an opponent's history to identify vulnerabilities and then formulating a direct counter-strategy to destroy that weakness.

Semi-Bluff: A bet or raise made with a hand that is not currently strong but has the potential to improve to a very strong hand (e.g., a flush or straight draw).

Set: Three of a kind made by holding a pocket pair and having one of those cards appear on the board.

Showdown: The final stage of a hand where all remaining players reveal their cards to determine the winner.

Slow Play: The tactic of playing a very strong hand passively (by checking or calling) to deceive opponents into thinking one's hand is weak, with the goal of inducing bluffs or calls on later streets. Also known as "trapping."

Solver: A type of poker software that calculates near-perfect GTO strategies for specific situations, typically through immense computational effort.

Stack-to-Pot Ratio (SPR): The ratio of the effective stack size to the size of the pot. This metric is crucial for planning commitment and strategy on the flop.

Statistical Profiling: The process of collecting and analysing data on an opponent's actions over many hands to build a detailed mathematical model of their playing style. This profile includes key metrics such as **VPiP** and **PFR**, which are then used to identify exploitable tendencies. This is the primary learning method used when cards are not revealed at a **showdown**.

Standard Raise: A pre-flop raise of a typical or expected size, usually around 3 times the **Big Blind**.

Steaming: A poker slang term for a player who is in a state of anger or frustration, often after a **Bad Beat**, which leads to irrational, overly aggressive decision-making. It is a form of **Tilt**.

Tight Aggressive (TAG): A standard player archetype characterised by a low **VPiP** (playing a narrow range of strong starting hands) and an aggressive post-flop style. It is a common style among solid, winning players.

**Tight-Medium Aggressive (TMAG):** A specific, granular classification used by **Patrick** to define a disciplined playing style. It is characterised by playing a relatively narrow range of strong starting hands, but playing them aggressively post-flop. It serves as one of the system's core baseline styles.

Tilt**:** A state of mental or emotional frustration, often triggered by a **Bad Beat**, in which a player adopts a sub-optimal strategy, resulting in irrational and often aggressive play. **Patrick**'s **HAA** module can simulate this state for strategic camouflage.

Turn: The second stage of post-flop play, in which the fourth community card is dealt, followed by a round of betting. It is also known as 'fourth street'.

Under the Gun (UTG): The position at the poker table that must act first in the pre-flop betting round.

VPiP (Voluntarily Put money in Pot): A key statistic that measures how frequently a player chooses to play a hand pre-flop (excluding forced blind bets). It is a primary indicator of how "tight" or "loose" a player is.

Whale: A player archetype, similar to a **Fish** or **Calling Station**, characterised by a very loose and often unpredictable style of play, making them a primary target for exploitation.

Yellow Line: The conventional term for the **All-in Adjusted** equity line on a Poker Tracker graph. While often used in a mistaken attempt to account for on-table luck , this paper posits that equating this metric with total variance is a common error. It represents only a tiny fraction of the total luck in a hand—the 'tip of the luck iceberg'—and can therefore be a misleading indicator of performance

## Appendix C: Detailed Analysis of Hand 6

This appendix provides the complete, granular data set used by the Ranges module during the Hand 6 case study. The following pages show the step-by-step analysis, including the specific Range Reshaping Templates (RETs), the resulting rS distributions, and the updated range grids for each stage of the hand.

The analysis uses two types of range grids. The first is the standard 169-square grid. The second, more detailed view is the **KuKulKan grid**, which displays all 1,326 possible two-card starting hand combinations. In both grid types, the colour of each square acts as a heatmap, where a darker shade indicates a higher probability that the opponent holds that specific hand.

## C.1: Pre-flop Range Assignment

The analysis begins with the initial range assigned to the 'Whale' following his call of Patrick's pre-flop raise. This range is visualized below in two formats: the standard 169-square grid and the more granular KuKulKan grid.

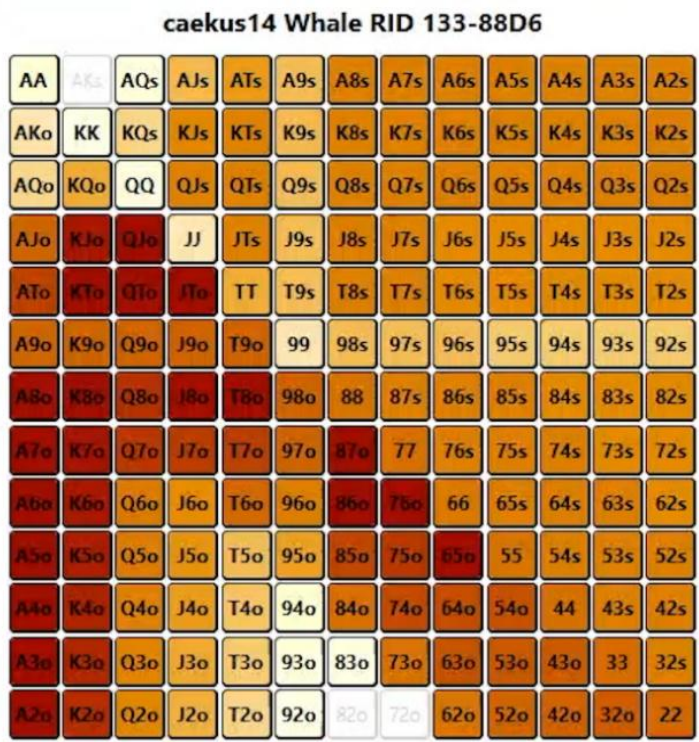

Figure C.1: The Standard Ranges Grid (169-square) showing the Whale's pre-flop range. *Heatmap: Darker shading indicates a higher probability.*

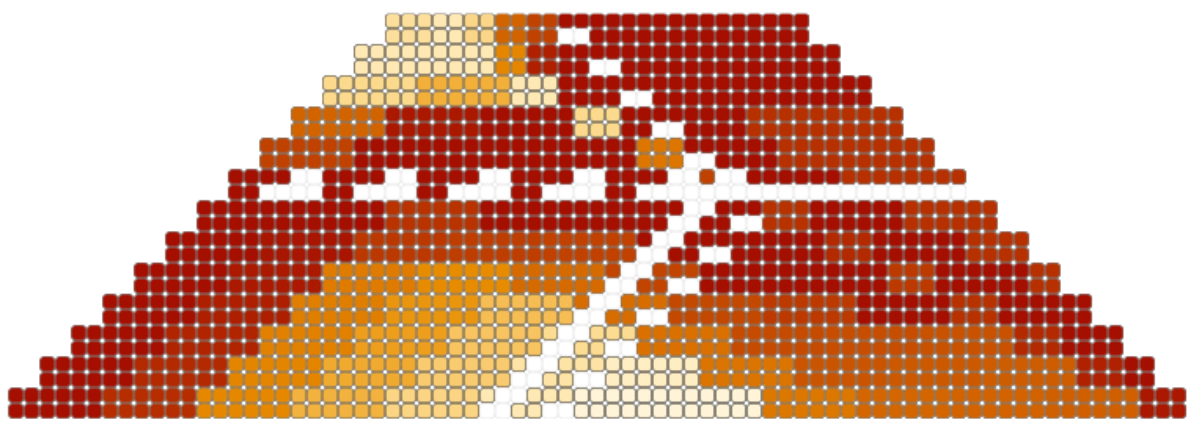

Figure C.2: The Whale's pre-flop range (KuKulKan grid). *Heatmap: Darker shading indicates a higher probability.*

## C.2: The Flop

After the 9♦5♠2♣ flop is dealt, the system performs an initial re-evaluation of the Whale's range. The resulting rS distribution and the updated range grids, processed with a flat RET to establish a new baseline, are shown below.

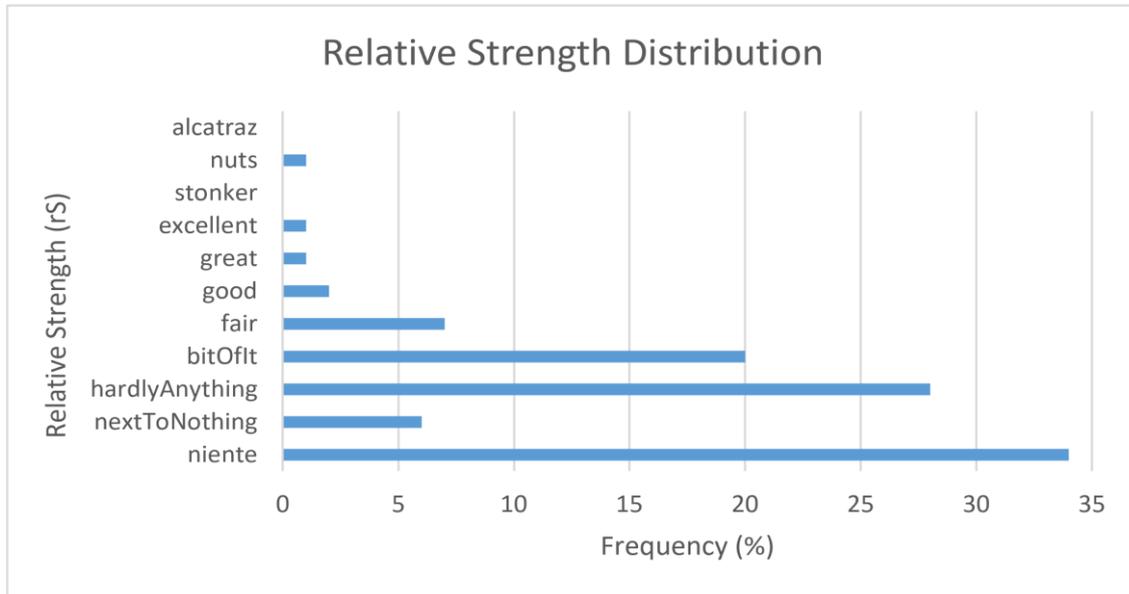

Figure C.3: The Whale's initial rS distribution on the flop (processed with a flat RET).

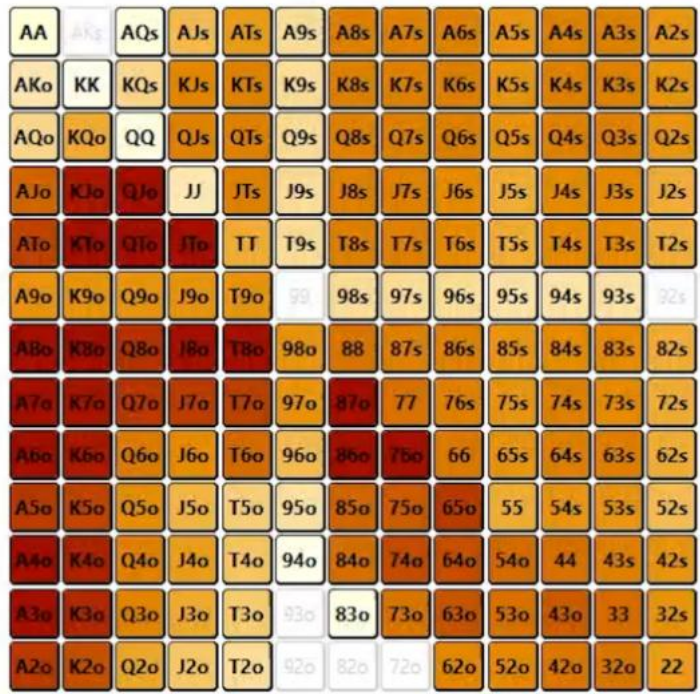

Figure C.4: The Standard Ranges Grid showing the Whale's updated hand range. *Heatmap: Darker shading indicates a higher probability.*

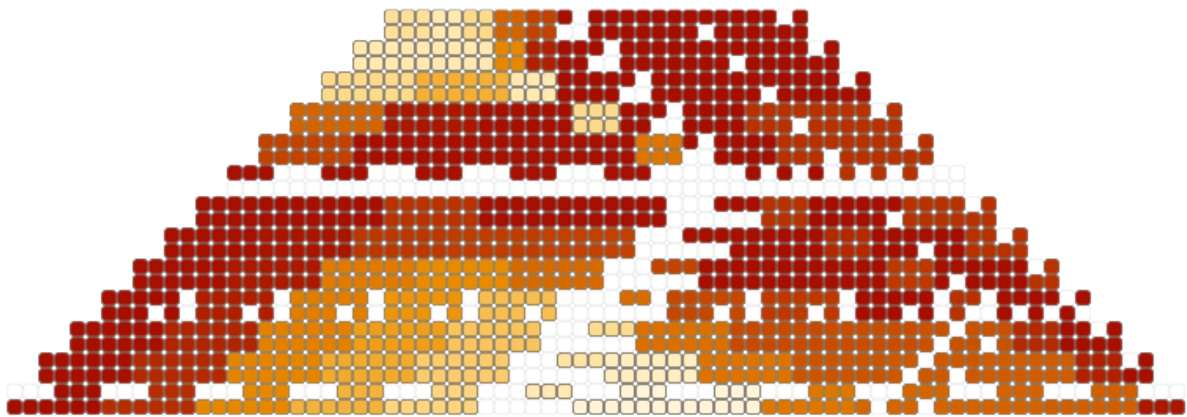

Figure C.5: The KuKulKan Grid showing the Whale's updated hand range. *Heatmap: Darker shading indicates a higher probability.*

## C.3: Flop Action: The Whale's Donk Bet

The first action on the flop is an unorthodox 'donk bet' from the Whale. The system processes this action by applying Range Reshaping Template (RET) 18. This template is heavily weighted toward strong made hands and key draws, and its application significantly reshapes the Whale's likely holdings, as shown in the updated distributions and range grids below.

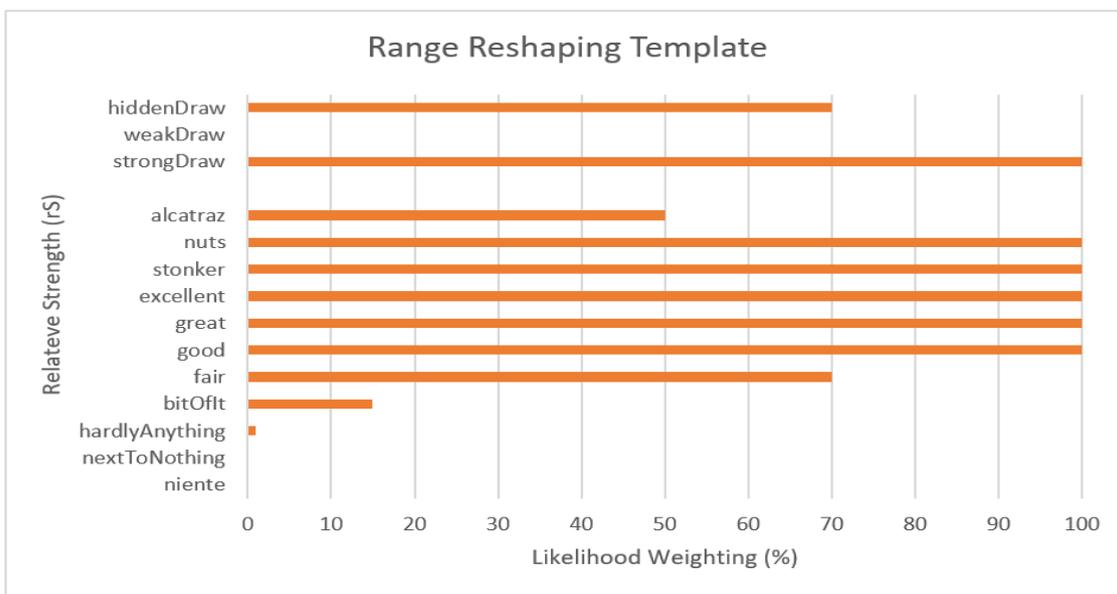

Figure C.6: RET 18: The reshaping template for the Whale's donk bet.

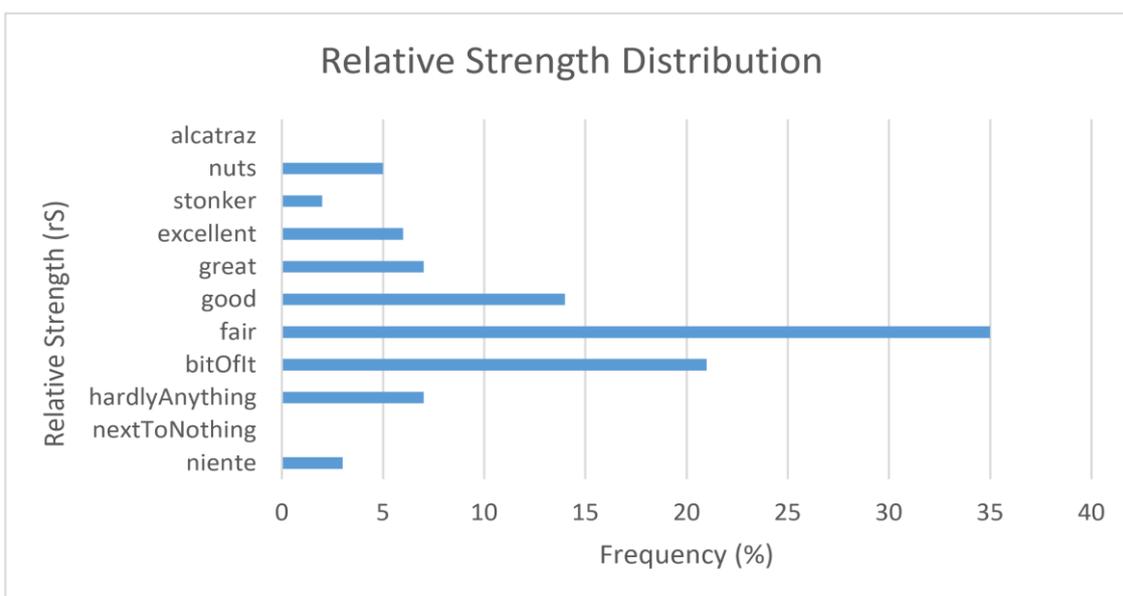

Figure C.7: The Whale's new rS distribution after his donk bet.

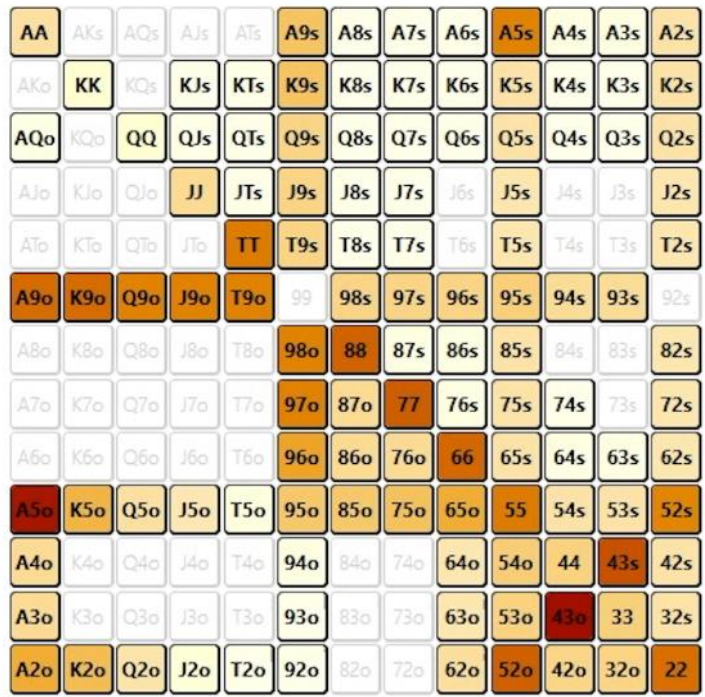

Figure C.8: The Standard Ranges Grid showing the Whale's range, reshaped by the donk bet. *Heatmap: Darker shading indicates a higher probability.*

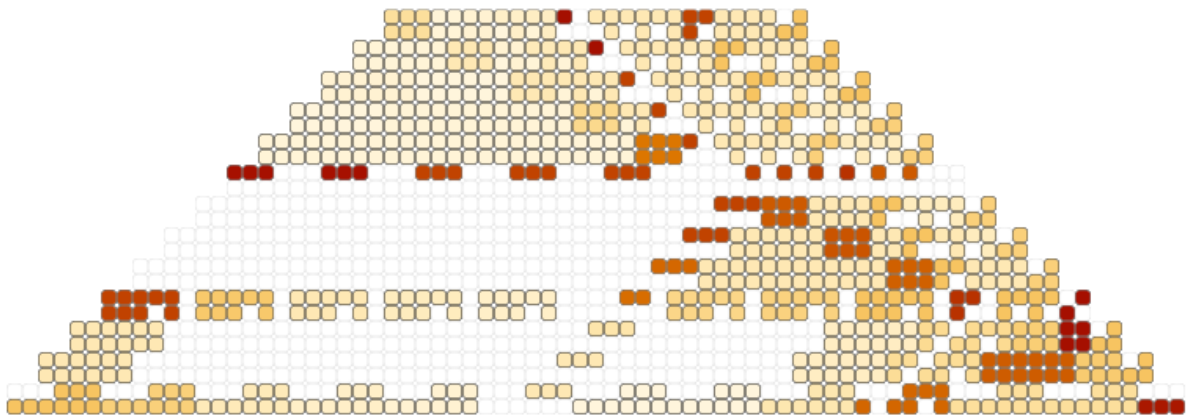

Figure C.9: The KuKulKan Grid showing the Whale's range, reshaped by the donk bet. *Heatmap: Darker shading indicates a higher probability.*

## C.4: Flop Action: The Whale's Call

Following Patrick's raise, the Whale calls. This action is processed with Range Reshaping Template (RET) 33. The application of this template further narrows the Whale's likely holdings, favouring medium-strength made hands and strong draws, as detailed in the figures below.

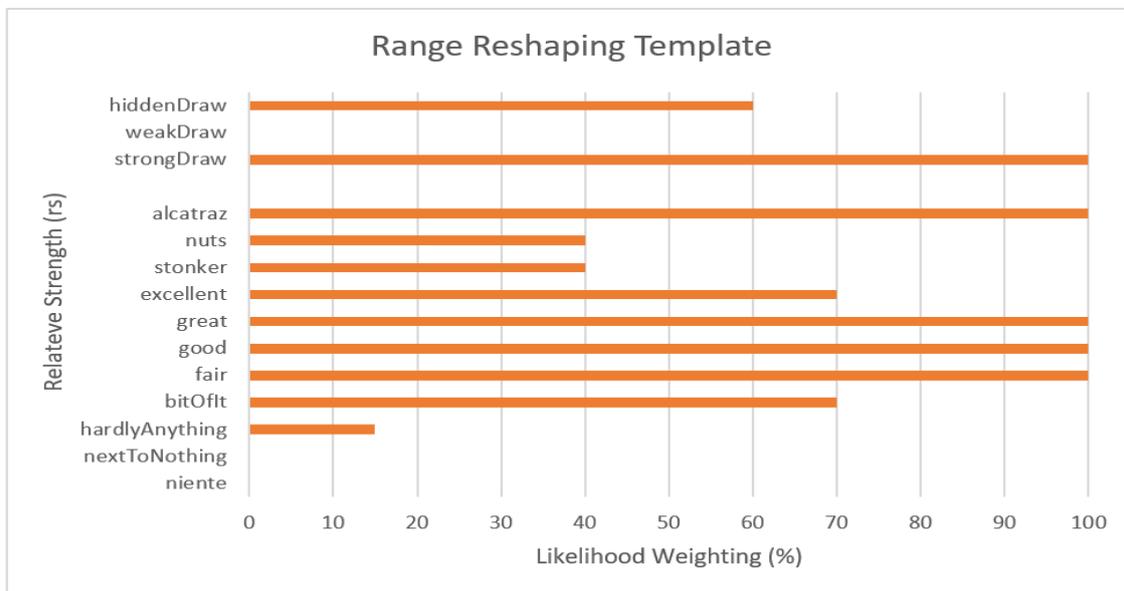

Figure C.10: RET 33: The reshaping template for the Whale's call.

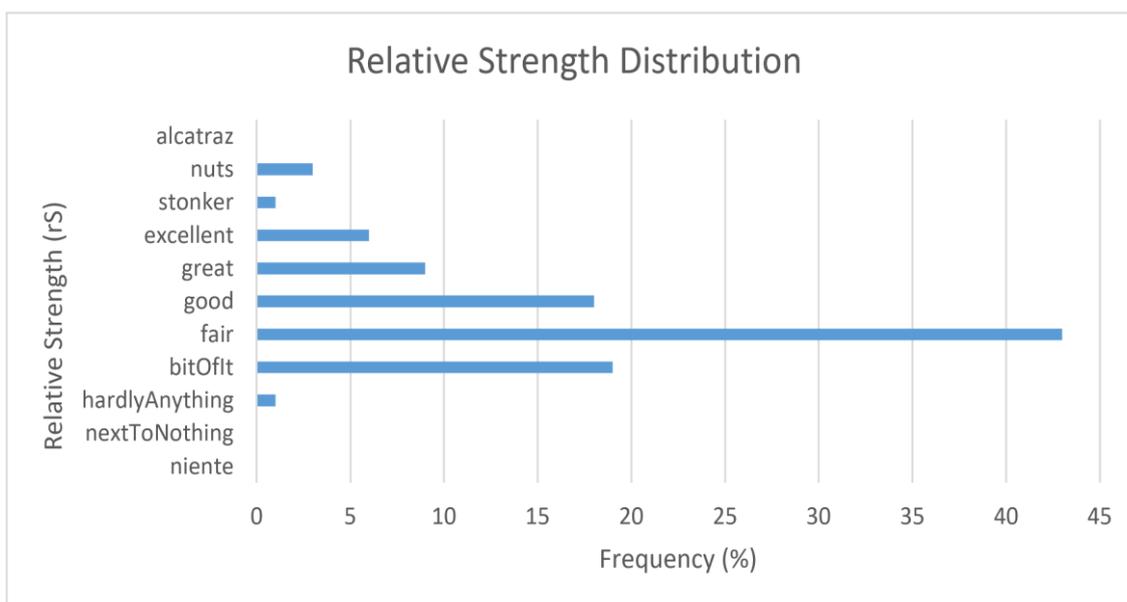

Figure C.11: The Whale's new rS distribution after his call.

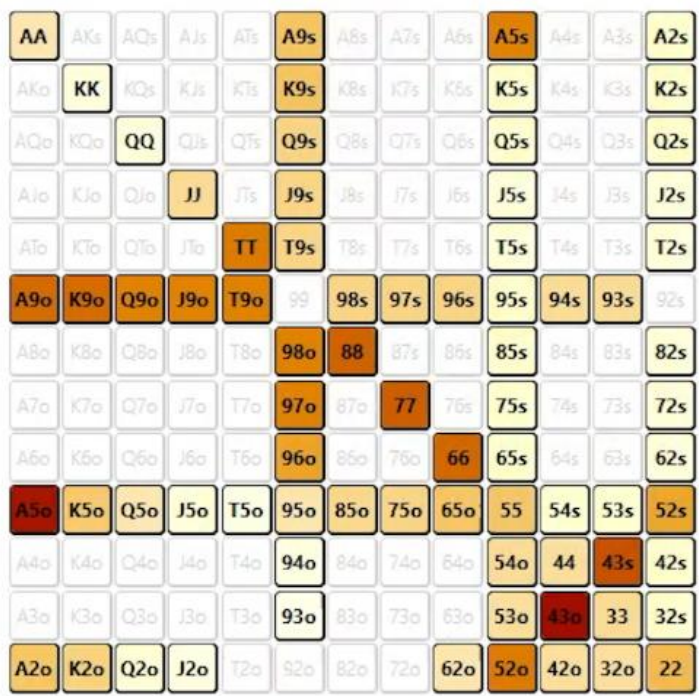

Figure C.12: The Standard Ranges Grid showing the Whale's range, reshaped by the call. *Heatmap: Darker shading indicates a higher probability.*

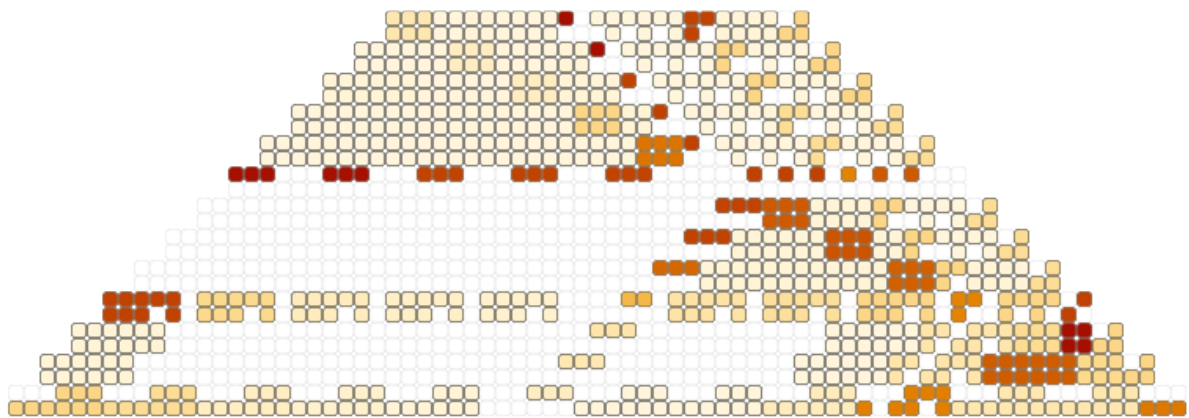

Figure C.13: The KuKulKan Grid showing the Whale's range, reshaped by the call. *Heatmap: Darker shading indicates a higher probability.*

## C.5: The Turn

The turn card is the 2♦. The figures below show the re-evaluation of the Whale's range after this card is dealt, but before any further betting action occurs. This analysis, processed with a flat RET, establishes a new baseline for the final round of betting.

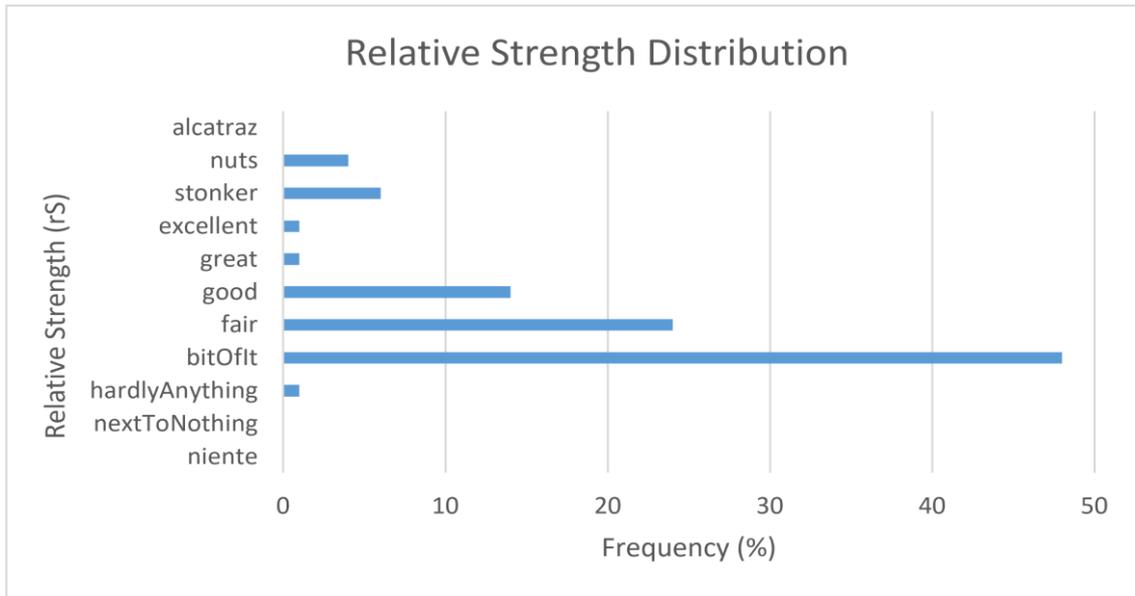

Figure C.14: The Whale's rS distribution after the turn card (processed with a flat RET).

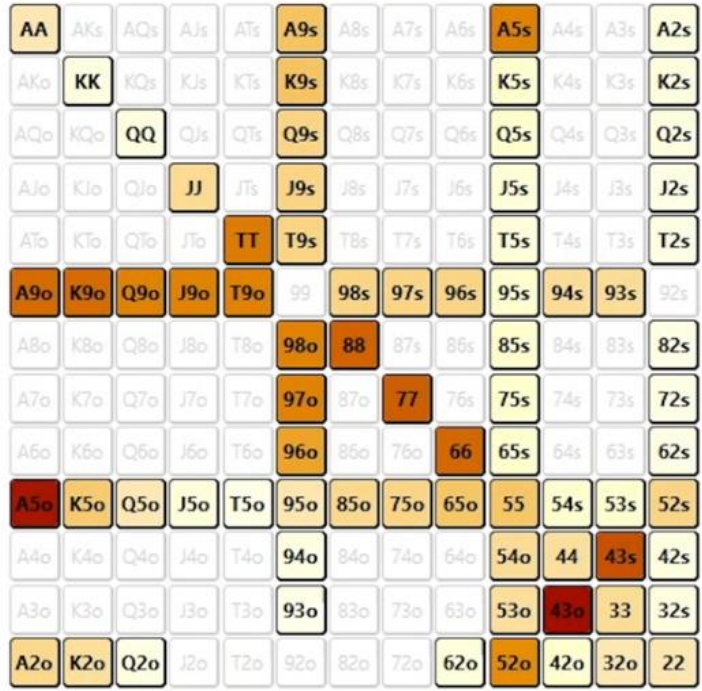

Figure C.15: The Standard Ranges Grid showing the Whale's updated hand range after the turn. *Heatmap: Darker shading indicates a higher probability.*

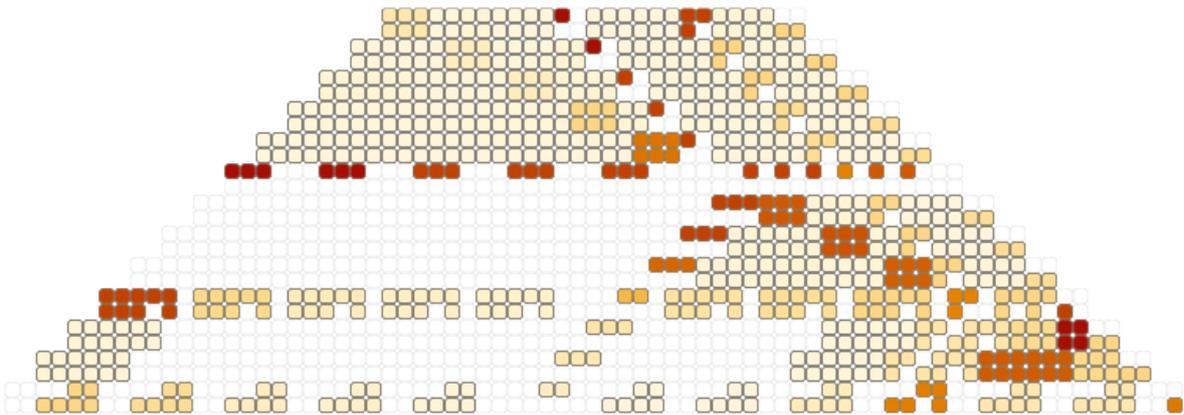

Figure C.16: The KuKulKan Grid showing the Whale's updated hand range after the turn. *Heatmap: Darker shading indicates a higher probability.*

## C.6: Turn Action: The Whale's All-In Bet

The final action on the turn is the Whale's all-in bet. This decisive move is analysed using Range Reshaping Template (RET) 73. The application of this template produces the system's final assessment of the Whale's hand range before Patrick calls, as detailed in the figures below.

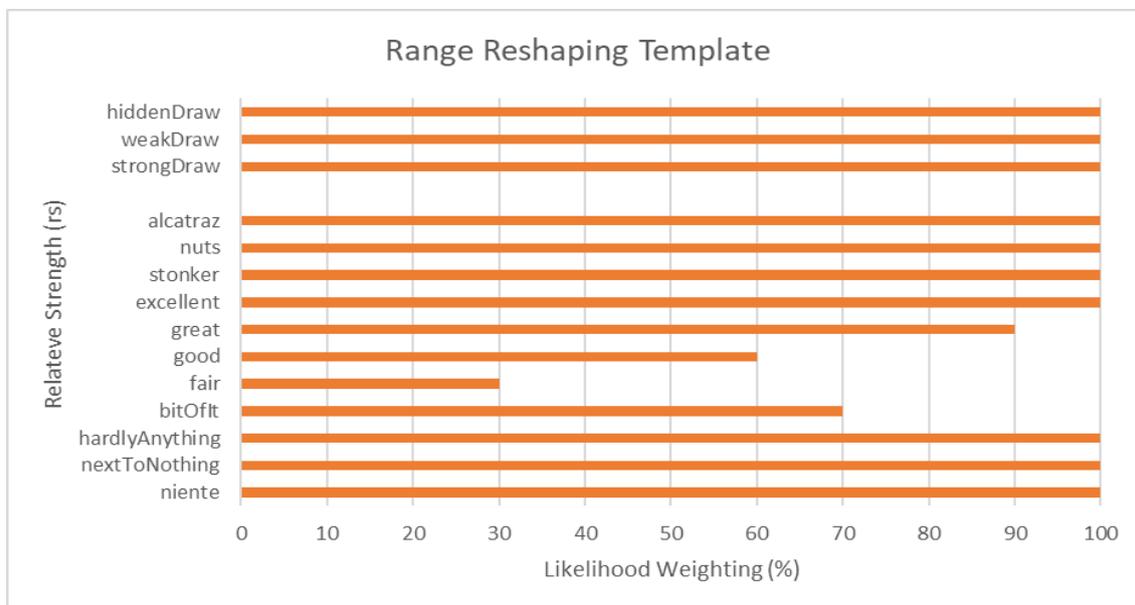

Figure C.17: RET 73: The reshaping template for the Whale's all-in bet.

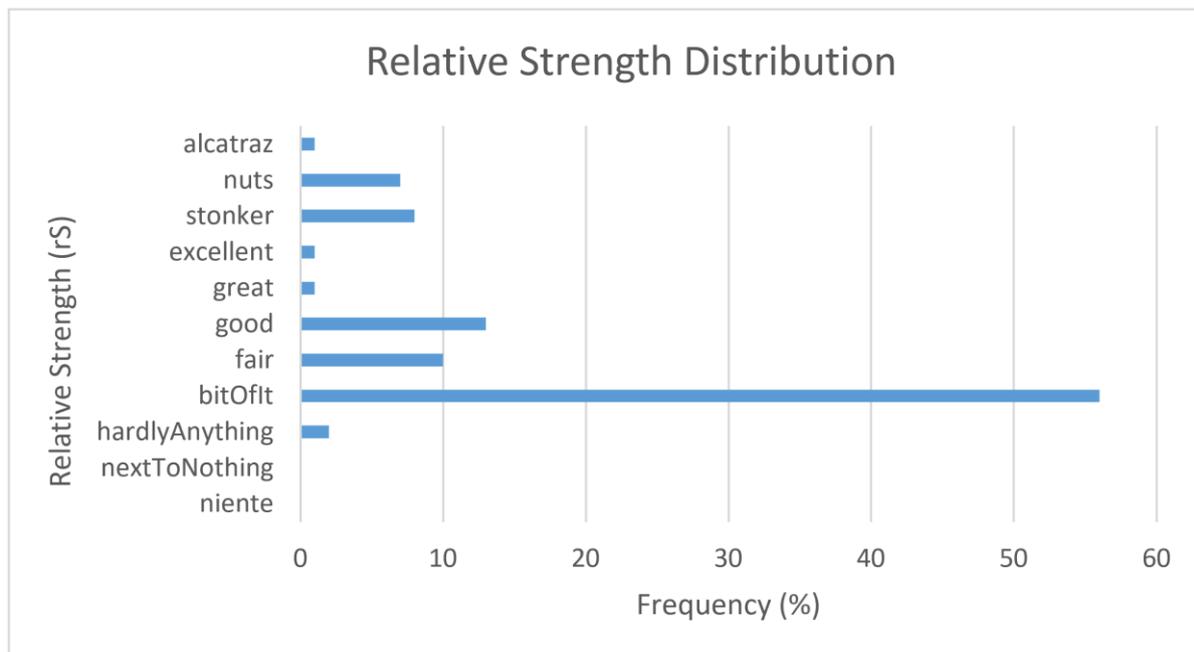

Figure C.18: The Whale's final rS distribution after his all-in.

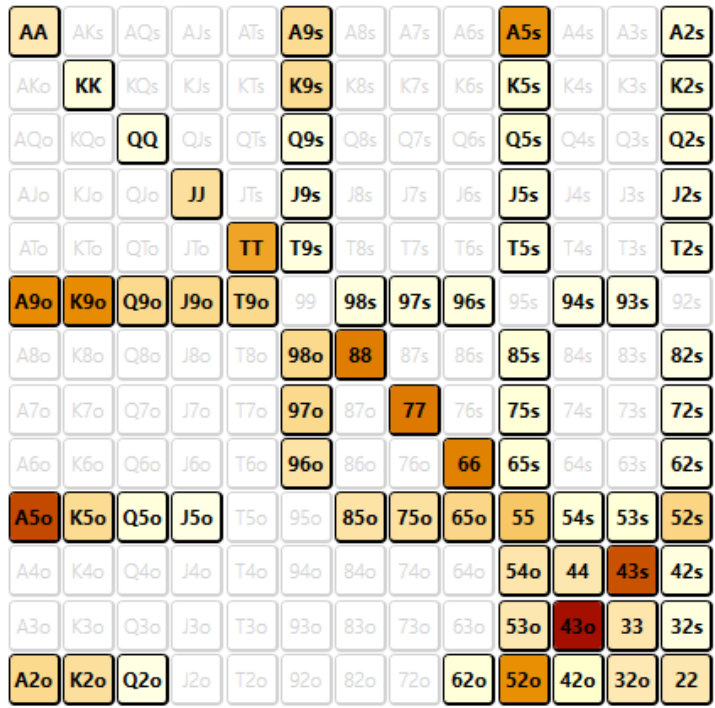

Figure C.19: The Standard Ranges Grid showing the Whale's final range. *Heatmap: Darker shading indicates a higher probability.*

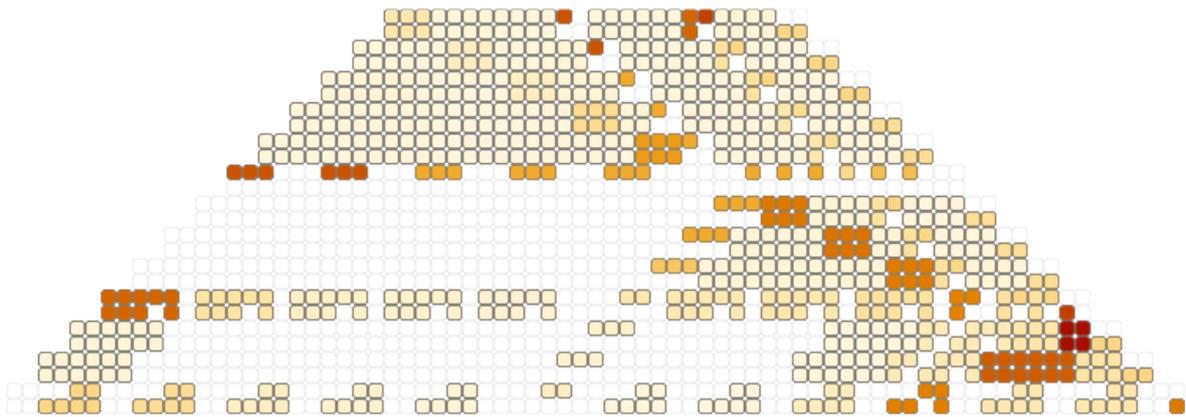

Figure C.20: The KuKulKan Grid showing the Whale's final range. *Heatmap: Darker shading indicates a higher probability.*

# Appendix D: Detailed Brain Architecture

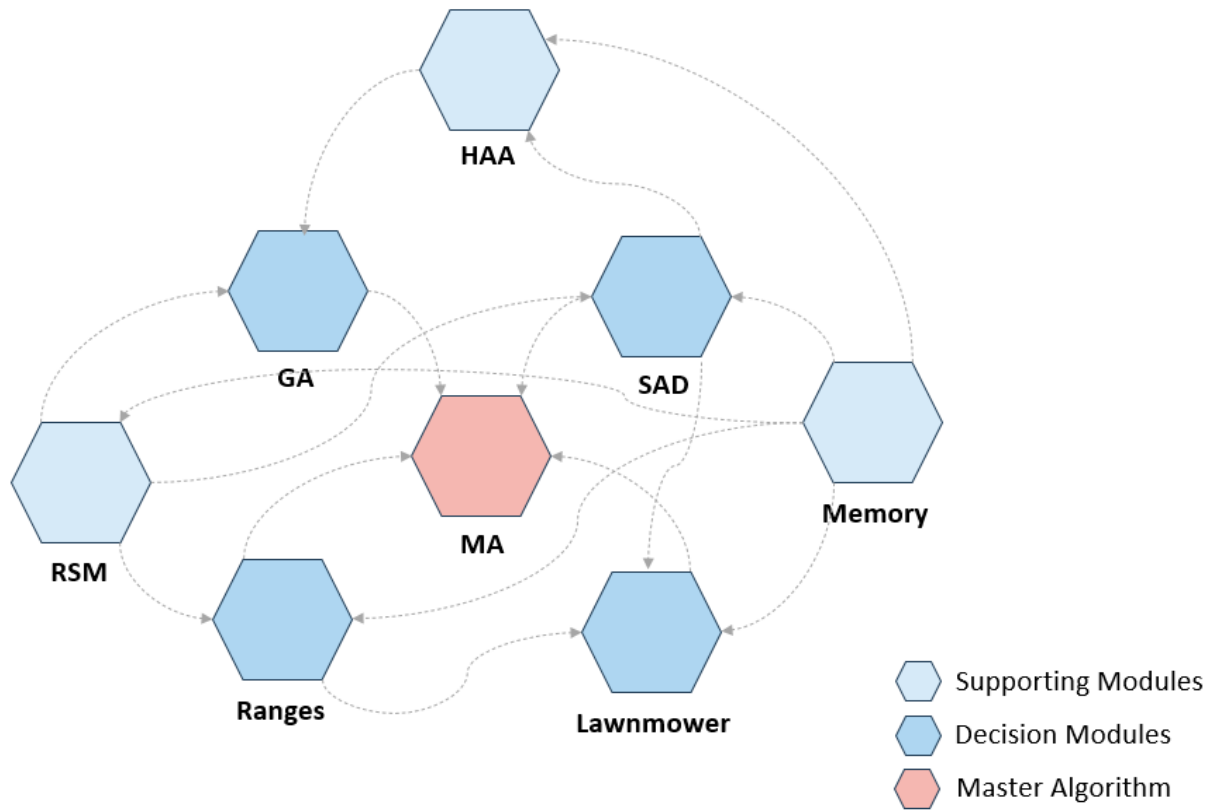

Figure D.1: A detailed schematic of the Brain module, illustrating the multi-directional flow of information between all components.